\documentclass[conference]{IEEEtran}
\IEEEoverridecommandlockouts

\usepackage[utf8]{inputenc}
\usepackage{times}
\usepackage{amsmath}
\usepackage{amssymb}
\usepackage{graphicx}
\usepackage{url}
\usepackage{color}
\usepackage{multirow}
\usepackage{array}
\usepackage{booktabs}
\usepackage{caption}
\usepackage{hhline}
\usepackage{colortbl} 
\usepackage{xcolor}   
\usepackage[hidelinks]{hyperref}
\hypersetup{
colorlinks=true,       
linkcolor=blue,          
citecolor=blue,        
urlcolor=black          
}

\usepackage{arydshln}

\newcolumntype{M}[1]{>{\centering\arraybackslash}m{#1}}
\newcolumntype{C}[1]{>{\centering\arraybackslash}m{#1}}

\usepackage{xurl} 
\begin{document}

\title{Towards Accurate UAV Image Perception: Guiding Vision-Language Models with Stronger Task Prompts}

\author{Mingning~Guo, Mengwei~Wu, Shaoxian~Li, Haifeng~Li, \textit{Senior Member,~IEEE}, and Chao~Tao, \textit{Member,~IEEE}
\thanks{This work was supported in part by the Natural Science Foundation of Hunan for Distinguished Young Scholars under Grant 2022JJ10072; in part by the National Natural Science Foundation of China under Grant 42471419 and Grant 42171376.}
\thanks{M.~Guo, M.~Wu, and S.~Li are with the School of Geosciences and Info-Physics, Central South University, Changsha 410083, China (e-mail: 225007012@csu.edu.cn, 235007016@csu.edu.cn, lishaoxian@csu.edu.cn). H.~Li and C.~Tao are with the School of Artificial Intelligence, Central South University, Changsha 410083, China (e-mail: lihaifeng@csu.edu.cn, kingtaochao@csu.edu.cn).}}

\maketitle
\begin{abstract}
UAV image perception based on Vision-Language Models (VLMs) aims to identify and localize targets through semantic and spatial descriptions in task prompts. However, small, densely distributed objects and cluttered backgrounds hinder these tasks, particularly when concise prompts provide insufficient guidance. Although task-specific fine-tuning can improve UAV image perception by strengthening image–text alignment, it typically requires large-scale annotated image–text pairs and model-specific parameter updates, increasing adaptation costs across models and tasks. To address this challenge, we propose \textit{AerialVP (Aerial Visual Perception)}, a task-adaptive, training-free prompt-enhancement agent that shifts the focus of adaptation from optimizing model parameters to enhancing task prompts. To enable task adaptation, AerialVP dynamically adjusts its enhancement plan based on the original prompt’s level of detail, selectively incorporating image-derived semantic attributes, spatial positions, and inter-object relationships. These complementary cues provide discriminative target descriptions and explicit spatial constraints, enabling VLMs to leverage their pretrained alignment and reasoning capabilities to improve task-conditioned image–text alignment and reduce target ambiguity. Specifically, AerialVP uses an LLM-based Task Engine to coordinate prompt-analysis and enhancement tools from a purpose-built modular Tool Repository for task analysis, visual information extraction, and prompt fusion. The enhanced prompt and original image are then provided to the unchanged backend VLM for prediction. Unlike model-centric fine-tuning, this prompt-centric strategy enables training-free adaptation across diverse backend VLMs, including open-source and proprietary models with different architectures. For systematic evaluation, we introduce \textit{AerialSense}, a unified UAV image perception benchmark covering Aerial Visual Grounding, Aerial Visual Reasoning, and Aerial Visual Question Answering across diverse scenes, resolutions, and imaging conditions. Extensive experiments demonstrate consistent performance improvements across all three tasks for the evaluated open-source and proprietary VLMs. AerialVP also achieves competitive performance with LoRA-based instruction fine-tuning under the evaluated settings, highlighting the effectiveness of prompt-centric adaptation for UAV image perception. Our work will be available at https://github.com/lostwolves/AerialVP.
\end{abstract}



\section{Introduction}
\IEEEPARstart{U}{AV} image perception extracts object locations, semantic attributes, and spatial relationships from aerial imagery \cite{cite1}, supporting autonomous navigation \cite{cite2} and intelligent decision-making \cite{cite3}. However, diverse viewpoints \cite{cite4}, large-scale variations \cite{cite5}, complex illumination \cite{cite6}, and cluttered backgrounds \cite{cite7} hinder accurate perception, particularly for small, densely distributed, and visually similar targets. Traditional vision-based methods rely on large-scale annotated datasets for supervised training to improve perception accuracy in specific scenarios \cite{cite8, cite9}. However, these datasets often fail to capture the diverse and dynamic environmental conditions encountered in real-world UAV applications, limiting the generalization of vision-based methods to unseen scenarios.

\begin{figure}[t]
\centering
\includegraphics[width=0.98\linewidth]{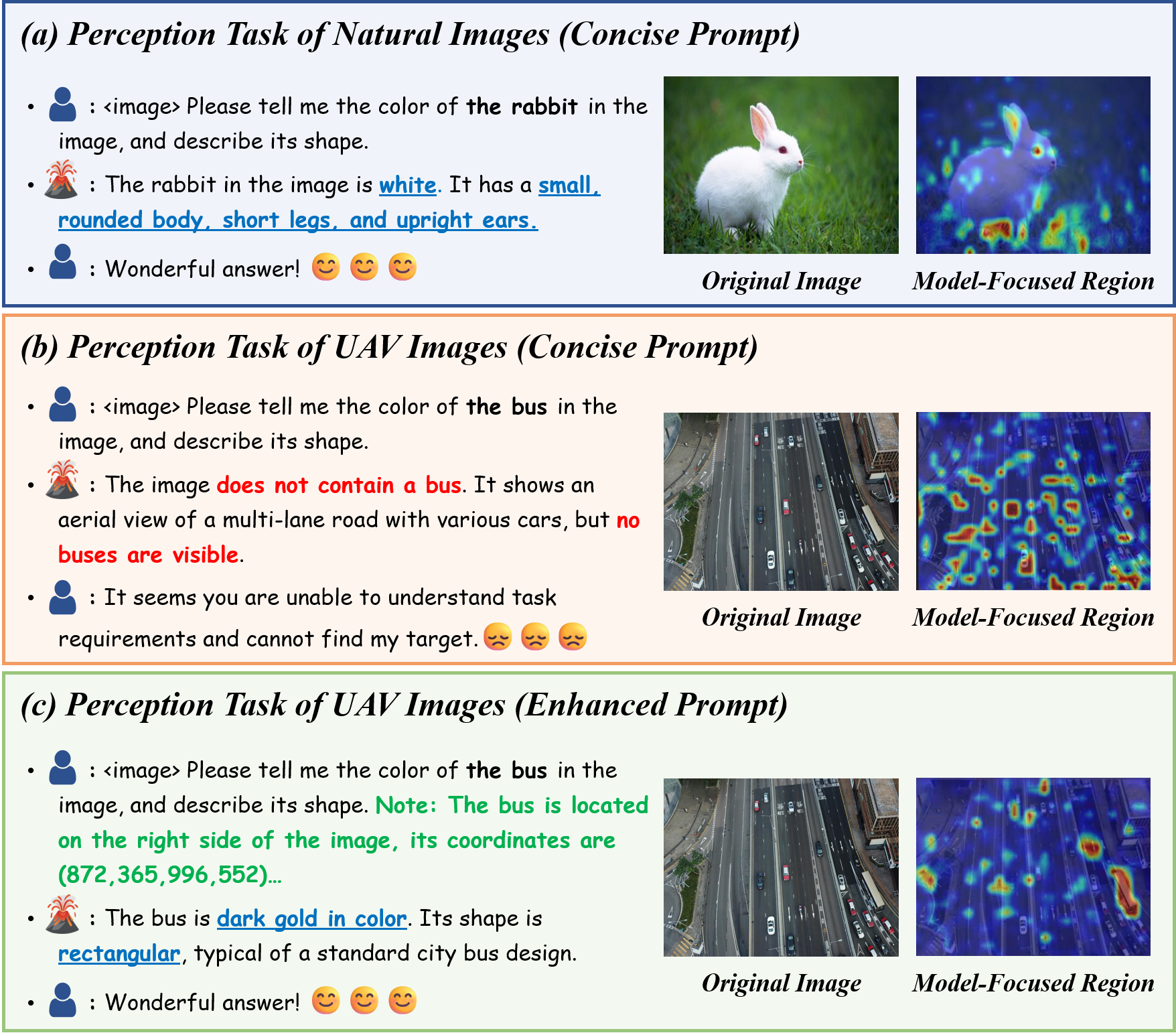}
\vspace{-1mm}
\caption{Impact of task prompt design on VLM perception under different visual environments. (a) In a simple scene, a concise prompt provides sufficient guidance for identifying the target. (b) In a complex UAV scene, insufficient task-relevant information in the concise prompt leads to ambiguous image-text alignment. (c) An enhanced prompt provides additional semantic and spatial guidance for more accurate perception.}
\vspace{-5mm} 
\label{fig:ques}
\end{figure}

With the rapid development of Vision-Language Models (VLMs), visual perception has benefited from stronger semantic understanding and generalization \cite{cite10, cite11, cite12}. By learning cross-modal alignment from large-scale image-text pairs \cite{cite13}, VLMs acquire generalizable representations \cite{cite14}, while language provides rich semantic priors for understanding diverse visual conditions \cite{cite15, cite16}. Despite these advances, applying VLMs to complex UAV imagery remains challenging. A fundamental prerequisite for effective multimodal perception is \textit{\textbf{precise semantic alignment between visual and textual representations within a shared representation space}} \cite{cite17, cite18}. This is because most VLMs perform cross-modal reasoning by projecting visual features into the language embedding space, where they interact and align with textual representations processed by the underlying Large Language Models (LLMs) \cite{cite19, cite20, cite21}. In many natural image scenarios, salient and spatially distinguishable objects can be effectively associated with concise textual descriptions \cite{cite22, cite23}. As shown in \textcolor{blue}{Fig. \ref{fig:ques} (a)}, the foreground target (\textit{“rabbit”}) is visually distinctive, allowing task-relevant visual tokens to be effectively associated with a concise textual prompt. In contrast, UAV imagery often contains densely distributed targets, large-scale variations, and cluttered backgrounds. As illustrated in \textcolor{blue}{Fig. \ref{fig:ques} (b)}, concise prompts may provide insufficient guidance for distinguishing task-relevant targets (\textit{“bus”}) from surrounding distractors, thereby impairing task-conditioned image–text alignment. These observations suggest that the mismatch between complex visual content and concise textual guidance hinders precise image–text alignment in the shared representation space, ultimately degrading VLM perception performance.

\begin{figure*}[ht]
\centering
\includegraphics[width=0.95\linewidth]{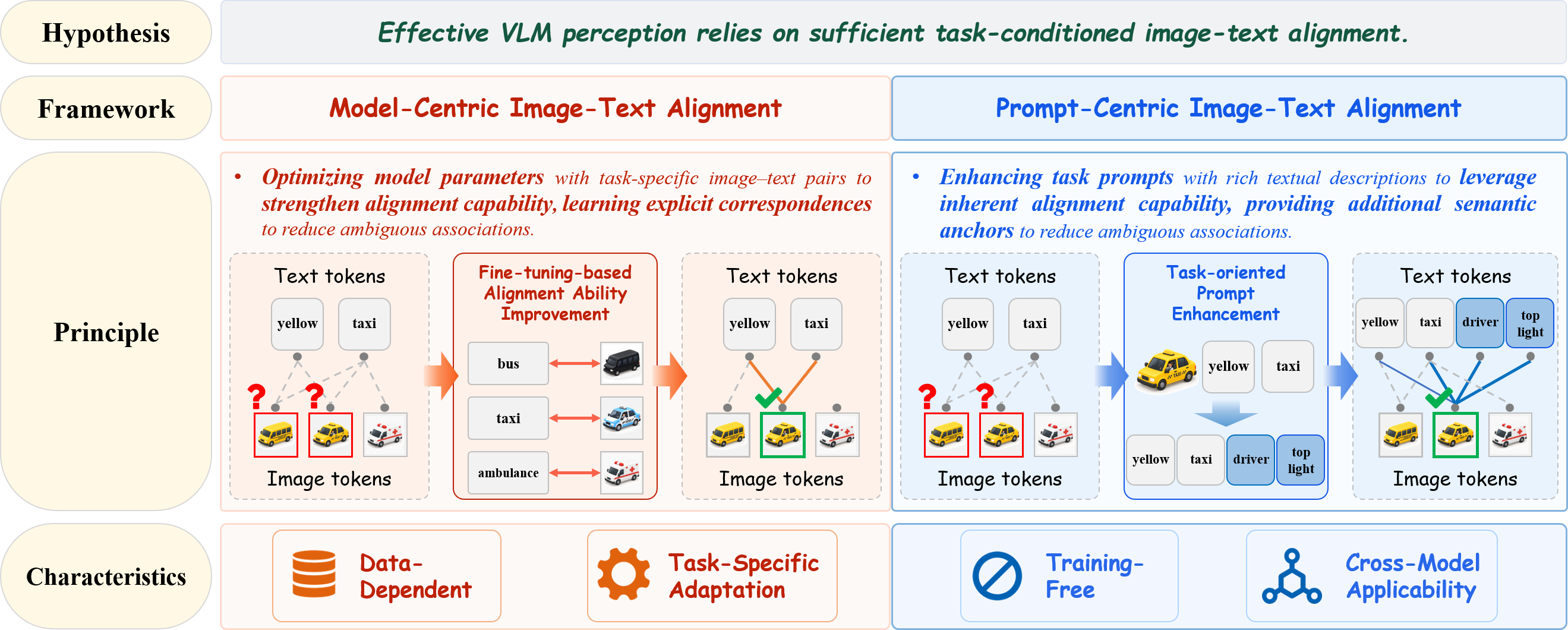}
\vspace{-1mm}
\caption{Comparison of Model-Centric and Prompt-Centric image-text alignment frameworks. The Model-Centric strategy learns task-specific alignment mappings through fine-tuning, strengthening correct image–text correspondences while suppressing ambiguous and spurious associations. In contrast, the Prompt-Centric strategy enriches textual descriptions with additional semantic anchors, mitigating alignment ambiguity and focusing image–text associations on task-relevant visual targets.}
\vspace{-3mm} 
\label{fig:compare}
\end{figure*}

Traditional fine-tuning approaches typically address this alignment difficulty by collecting task-specific image–text pairs and optimizing VLM parameters to strengthen image–text alignment \cite{cite24, cite25, cite26}. We refer to this paradigm as the \textit{\textbf{Model-Centric Framework}}. As illustrated on the left side of \textcolor{blue}{Fig. \ref{fig:compare}}, this paradigm learns explicit image–text correspondences through parameter optimization, strengthening the model’s ability to associate textual descriptions with the correct visual targets while reducing ambiguous associations. However, its effectiveness relies heavily on large-scale, task-specific annotated image–text data, and parameter adaptation is typically required for different models or tasks. These requirements increase adaptation costs and limit flexibility across diverse UAV perception scenarios. Moreover, parameter-level alignment optimization does not directly enrich the task information available at inference time. When user prompts are substantially simpler than the rich textual descriptions used during task-specific fine-tuning, the available semantic and spatial references may remain insufficient for precise task-conditioned image–text alignment.

Studies of human visual cognition provide a useful perspective on this problem. During human visual search, the top-down visual attention mechanism guides attention toward candidate regions that match target-defining features, while irrelevant candidates are progressively suppressed \cite{cite27, cite28, cite29}. Linguistic information can further interact with visual context to conduct refined screening and localization of candidate regions, thereby biasing attention toward the correct objects that are visually or semantically compatible \cite{cite30}. For example, when searching for \textit{“the apple with red-and-white stripes”} among a basket of fruits, humans can first use the category and appearance cues to narrow the search to a set of plausible candidates, but may still be unable to identify the intended apple unambiguously. If additional guidance is provided, such as \textit{“located in the upper-right corner of the basket, with two green leaves attached to the stem,”} the candidate set can be further constrained, enabling precise identification of the target. These findings suggest that additional linguistic guidance provides discriminative semantic anchors, enabling more precise associations between target concepts and visual candidates while reducing perceptual ambiguity.

Inspired by these observations, we shift the focus from model-centric alignment optimization to task-oriented prompt design. As illustrated in \textcolor{blue}{Fig. \ref{fig:ques} (c)}, we argue that \textit{\textbf{enhancing textual prompts with task-relevant visual information can strengthen task-conditioned image–text alignment during inference without modifying the backend VLM parameters.}} We refer to this paradigm as the \textit{\textbf{Prompt-Centric Framework}}. As illustrated in \textcolor{blue}{Fig. \ref{fig:compare}}, rather than improving the model’s alignment capability through parameter optimization, the Prompt-Centric Framework leverages the inherent alignment capability of the pretrained VLM by providing richer task-oriented textual descriptions. These enriched descriptions introduce additional semantic anchors, enabling more precise associations between task descriptions and the correct visual targets while reducing interference from ambiguous candidates. Compared with the Model-Centric Framework, this strategy performs adaptation at the prompt level in a training-free manner, enabling flexible deployment across different VLM architectures.

However, applying this paradigm to VLM-based UAV perception introduces two key challenges. First, the enhanced information should be task-relevant rather than merely increasing prompt complexity, as irrelevant cues may introduce noise and distract the model from the intended target. Unlike humans, who can leverage prior knowledge and contextual understanding to select discriminative information for a specific task, VLMs require explicit guidance to determine which visual cues should be incorporated for effective alignment. Second, the auxiliary information should be expressed through concepts that can be reliably understood and aligned by the VLM, avoiding additional ambiguity introduced by unfamiliar descriptions. Humans develop robust visual–language associations through long-term learning and can flexibly interpret diverse descriptions. In contrast, VLMs may struggle with rare, domain-specific, or weakly grounded concepts that are insufficiently represented during pretraining, potentially leading to unreliable associations and additional alignment ambiguity.

To address these challenges, we propose \textit{\textbf{AerialVP} (\textbf{Aerial} \textbf{V}isual \textbf{P}erception)}, a task-adaptive prompt-enhancement agent for UAV image perception. AerialVP consists of an LLM-based Task Engine and a modular Tool Repository, and performs prompt enhancement through three stages: \textbf{Task Prompt Analysis}, \textbf{Tool Selection}, and \textbf{Enhanced Prompt Generation}. To enable task-adaptive enhancement, the Task Engine first analyzes the original prompt to identify the task type, relevant objects, and required enhancement dimensions, thereby determining what information should be incorporated. To ensure semantic compatibility with pretrained VLMs, AerialVP employs specialized tools in the Tool Repository to extract visual cues represented by fundamental concepts, including object attributes, spatial positions, and inter-object relationships. These concept-level descriptions are naturally grounded in the semantic space learned by VLMs, enabling reliable image–text alignment while avoiding ambiguity introduced by unfamiliar descriptions. Finally, the extracted information is integrated with the original prompt to construct an enhanced task prompt, which is provided together with the original UAV image to the backend VLM for final inference. In this way, AerialVP realizes a flexible inference-time adaptation mechanism that strengthens task guidance while preserving the backend VLM.

For systematic evaluation, we further construct \textit{\textbf{AerialSense}}, a multi-task benchmark for UAV image perception by integrating and reorganizing multiple publicly available UAV and aerial datasets. AerialSense contains 7,119 UAV images and 53,374 task samples covering three representative tasks: Aerial Visual Grounding, Aerial Visual Reasoning, and Aerial Visual Question Answering. The benchmark includes diverse target categories, image resolutions, illumination conditions, as well as urban and natural environments, enabling evaluation of VLM perception under a broad range of UAV imaging conditions. Extensive experiments on AerialSense demonstrate that AerialVP consistently improves the performance of both open-source and proprietary VLMs across all three perception tasks. Qualitative analyses further show that enhanced textual prompts guide VLMs toward task-relevant visual regions, thereby strengthening image-text alignment during inference. Comparison with LoRA-based instruction fine-tuning shows that the proposed Prompt-Centric Framework achieves competitive performance under the evaluated settings.

The main contributions of this paper are summarized as follows:
\begin{itemize}
\item This study is the first to systematically analyze the challenges of VLM-based UAV image perception from the perspective of image–text alignment. We reveal that alignment ambiguity stems from two factors: on the visual side, dense and similar UAV targets cause multiple candidate regions to correspond to the same or closely related semantic concepts learned from natural-image pretraining; on the textual side, concise task prompts provide insufficient semantic constraints for candidate disambiguation. This provides new perspectives and directions for addressing image–text alignment challenges in complex UAV environments.
\item This study proposes a task-adaptive, training-free Prompt-Centric paradigm for improving task-conditioned image–text alignment. Instead of optimizing model parameters, it enriches task prompts with more informative textual references for image–text alignment. We instantiate this paradigm as AerialVP, an agent for UAV image perception that dynamically determines task-specific information needs and selectively incorporates image-derived semantic attributes, spatial positions, and inter-object relationships into the task prompt as discriminative alignment cues. AerialVP supports both open-source and proprietary VLMs without additional training and achieves consistent improvements across all evaluated models and tasks, with competitive performance against LoRA-based instruction fine-tuning.
\item This study develops AerialSense, a unified multi-task benchmark for systematically evaluating VLM-based UAV image perception. AerialSense jointly covers Aerial Visual Grounding, Aerial Visual Reasoning, and Aerial Visual Question Answering across diverse target categories, scenes, image resolutions, and imaging conditions. This benchmark enables comprehensive and standardized evaluation of VLM localization, semantic understanding, and reasoning capabilities in complex UAV scenarios.
\end{itemize}


\section{Related Work}
\subsection{Image-Text Alignment for Visual Perception}
The effectiveness of VLMs fundamentally depends on semantic alignment between visual and textual representations. Traditional alignment-based approaches typically establish this alignment using large-scale image-text pairs and further improve it by optimizing model parameters with task-specific data \cite{cite31, cite32, cite33}. Following this direction, extensive studies have explored more accurate and robust image-text alignment. For example, FG-CLIP \cite{cite34} improves fine-grained region-text correspondence through large-scale region-level annotations and detailed textual descriptions, while DC-CLIP \cite{cite35} strengthens multilingual alignment using bilingual image-text data and feature distillation. FLAVARS \cite{cite36} further incorporates geospatial context and positional encoding to support spatially aware representation learning, and DAPT \cite{cite37} enhances cross-modal interaction through foreground-background decoupling.

These studies demonstrate the effectiveness of parameter optimization for improving image-text alignment. However, its application to UAV perception is limited by the scarcity of diverse UAV image-text data and the mismatch between rich training descriptions and concise inference prompts. In contrast, our work adaptively enhances task prompts with image-derived semantic and spatial information according to task requirements, providing stronger task-specific guidance for accurate image perception.

\subsection{Prompt Enhancement in Multimodal Learning}
Recent studies have shown that task prompts play an important role in guiding VLM inference \cite{cite38, cite39, cite40}. Well-designed prompts can provide additional semantic or spatial context, reduce ambiguity, and guide models toward task-relevant information. Existing prompt-enhancement studies mainly explore textual prompt augmentation, visual prompt augmentation, and domain-specific prompt design. For textual prompt enhancement, AdaptCLIPZS \cite{cite41} introduces LLM-generated category descriptions to provide richer semantic information and improve zero-shot recognition. For visual prompt enhancement, FGVP \cite{cite42} and FGVTP \cite{cite43}incorporate fine-grained visual cues derived from segmentation or image-text consistency to improve visual grounding and part-level recognition. Prompt-guided strategies have also been explored in domain-specific applications, including robotic control \cite{cite44} and medical visual reasoning \cite{cite45}, demonstrating that additional task context can improve multimodal inference without requiring parameter updates to the target model.

However, most existing methods are task-specific or rely on predefined enhancement strategies, limiting their adaptability to diverse perception requirements. This limitation is particularly evident in UAV scenarios, where different tasks may require different combinations of semantic and spatial information. In contrast, AerialVP analyzes task requirements and selectively incorporates image-derived semantic and spatial cues for task-adaptive prompt enhancement.

\subsection{Tool-Augmented Multimodal Agents}
With the development of LLMs, tool-augmented agents increasingly employ LLMs as high-level controllers to decompose tasks and coordinate external vision models, language models, or computational tools \cite{cite46, cite47}. This paradigm has been extended to multimodal understanding and embodied applications, where LLMs perform task planning and tool selection while specialized modules execute perception operations \cite{cite48, cite49}. Standardized interfaces and tool-interaction protocols, such as the Model Context Protocol (MCP) \cite{cite50}, further facilitate the integration and invocation of heterogeneous tools.

AerialVP follows this tool-augmented paradigm but focuses specifically on task prompt enhancement for UAV perception. Its Tool Repository separates Prompt Analysis Tools for identifying task requirements from Prompt Enhancement Tools for extracting semantic and spatial information. The LLM-based Task Engine coordinates these tools and integrates their outputs into an enhanced prompt, while the final prediction remains generated by the backend VLM. This design enables task-adaptive and training-free prompt enhancement across different UAV perception tasks.


\begin{figure*}[ht]
\centering
\includegraphics[width=0.9\linewidth]{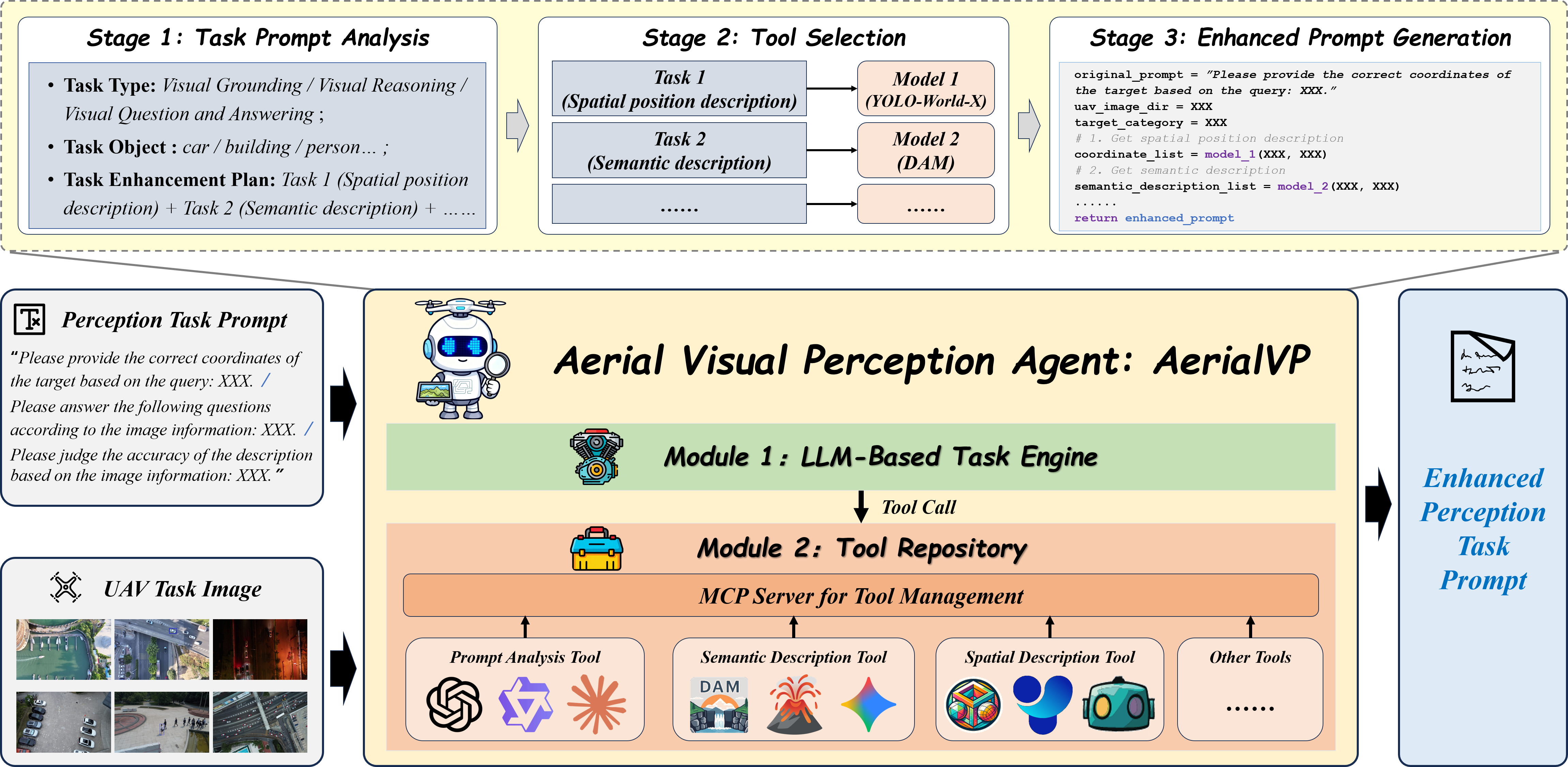}
\vspace{-1mm}
\caption{Overall architecture of AerialVP for task-adaptive prompt enhancement, comprising an LLM-based Task Engine and a modular Tool Repository.}
\vspace{-1mm} 
\label{fig:overview}
\end{figure*}

\section{AerialVP: A Prompt-Enhancement Agent for UAV Image Perception Tasks}
This section first formulates the VLM-based UAV image perception paradigms. Then we present the overall architecture of AerialVP, which consists of an LLM-based Task Engine and a modular Tool Repository. Building on this architecture, we further describe the overall prompt-enhancement workflow of AerialVP and the main functions of each stage.
\subsection{Problem Formulation}
Given a UAV image $I$ and an original task prompt $P_0$, a pretrained VLM directly generates a task-specific prediction as:
\begin{equation}
R = f_{\text{VLM}}(P_0, I)
\end{equation}
where $f_{\text{VLM}}$ denotes the pretrained VLM, and $R$ represents the final prediction. Depending on the task, $R$ may correspond to a target location in Visual Grounding (VG), a binary decision in Visual Reasoning (VR), or an answer option in Visual Question Answering (VQA).

In conventional VLM-based perception, the pretrained VLM relies on its existing image–text alignment capability to associate the semantic and spatial descriptions in $P_0$ with the relevant visual content in $I$. However, complex UAV imagery often contains small, densely distributed, and visually similar targets, making precise task-conditioned image--text alignment difficult when the original prompt provides limited discriminative information.

A conventional Model-Centric Framework addresses this difficulty by adapting the VLM parameters with task-specific image--text data. After fine-tuning, perception is performed using the same original prompt and UAV image:
\begin{equation}
    R_{\text{MC}} = f_{\text{VLM}}^{\text{ft}}(P_0, I)
\end{equation}
where $f_{\text{VLM}}^{\text{ft}}$ denotes the VLM after task-specific fine-tuning and $R_{\text{MC}}$ denotes the resulting prediction. In this paradigm, adaptation occurs on the model side: the input remains unchanged, while parameter optimization strengthens the model's capability to align task-relevant textual descriptions with visual content.

In contrast, the Prompt-Centric Framework performs adaptation on the input side while keeping the pretrained VLM unchanged. Given $P_0$ and $I$, a generic prompt-enhancement method $A(\cdot)$ first constructs an enhanced task prompt $P_e$; the final prediction is then obtained using the original pretrained VLM:
\begin{equation}
    P_e = A(P_0, I)
\end{equation}
\begin{equation}
    R_{\text{PC}} = f_{\text{VLM}}(P_e, I)
\label{eq:enhanced}
\end{equation}
where $P_e$ denotes the enhanced prompt containing additional task-relevant information derived from the UAV image; $R_{\text{PC}}$ denotes the prediction under the Prompt-Centric Framework. Unlike Model-Centric adaptation, this paradigm does not modify the backend VLM parameters. Instead, it strengthens the textual references available for task-conditioned image--text alignment by enriching the inference context before prediction.

Accordingly, the key distinction between the two adaptation paradigms lies in where adaptation occurs. The Model-Centric Framework improves alignment capability through parameter optimization while retaining the original input $(P_0, I)$, whereas the Prompt-Centric Framework retains the original VLM and improves the information provided for alignment by transforming $P_0$ into $P_e$. As a concrete implementation of the Prompt-Centric Framework, AerialVP realizes this input-side adaptation through task-adaptive prompt enhancement. The overall architecture and prompt-enhancement workflow of AerialVP are introduced in the following section.

\begin{figure*}[ht]
	\centering
	\includegraphics[width=0.85\linewidth]{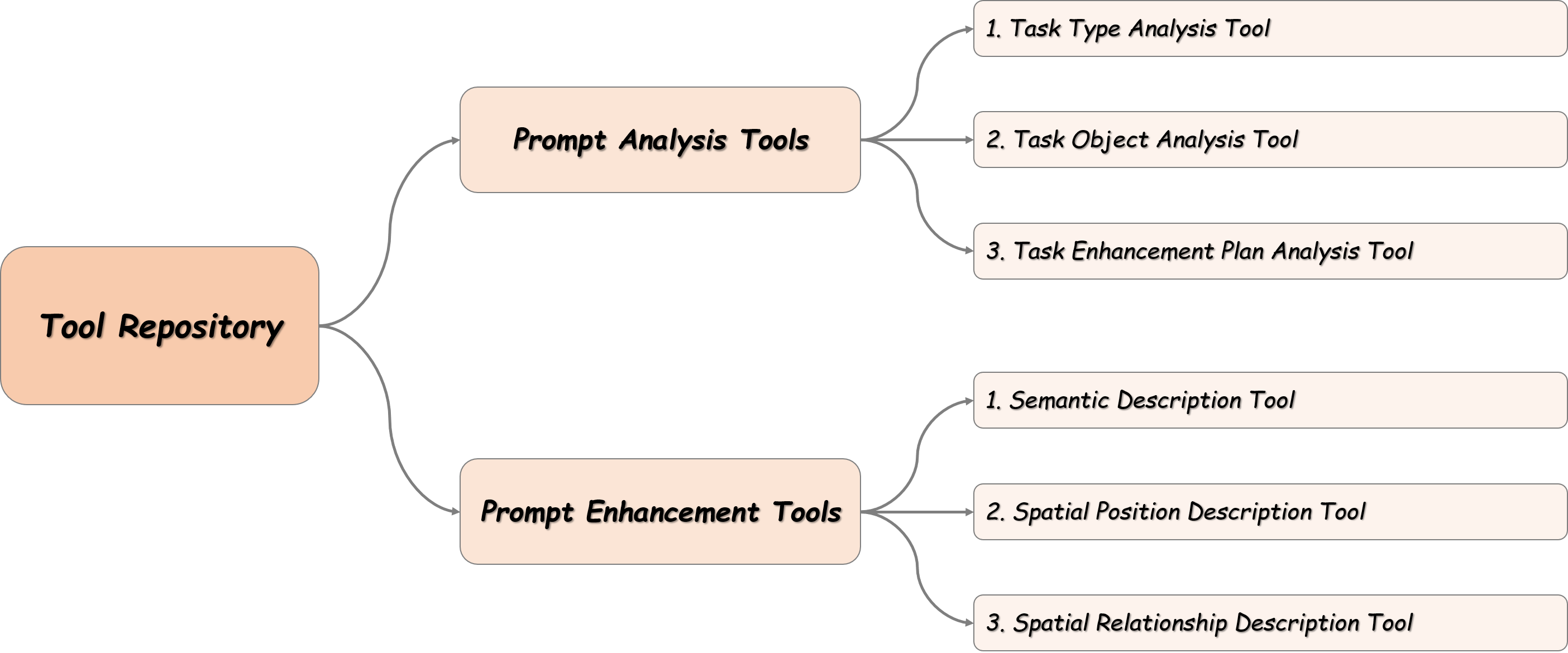}
	\vspace{-1mm}
	\caption{Structure of the Tool Repository in the AerialVP framework. The repository consists of Prompt Analysis Tools for task understanding and Prompt Enhancement Tools for extracting semantic and spatial information from UAV imagery.}
	\vspace{-2mm} 
	\label{fig:tool_tree}
\end{figure*}

\subsection{Core Components of AerialVP}
As illustrated in \textcolor{blue}{Fig. \ref{fig:overview}}, AerialVP consists of two core components: \textbf{an LLM-based Task Engine $E$} and \textbf{a modular Tool Repository $M$}. The Task Engine serves as the high-level controller responsible for task management and workflow planning, while the Tool Repository provides functional tools for task analysis and task-relevant visual information extraction. Through their collaboration, AerialVP transforms concise task prompts into informative prompts for accurate UAV image perception.
\subsubsection{Module 1. LLM-Based Task Engine}\hfill\\
\indent The Task Engine $E$ serves as the high-level controller of AerialVP. We adopt GPT-4.1 as its backbone due to its strong instruction-following and reasoning capabilities. Rather than directly performing visual perception, the Task Engine interprets the current execution state, determines the required operation, and translates high-level task requirements into executable instructions.

At each execution stage, the Task Engine operates on three types of information. First, it receives the task context, including the original task prompt $P_0$ and the intermediate results accumulated from preceding stages. Second, it accesses the tool context provided by the Tool Repository $M$, including the names, functional descriptions, input arguments, and output formats of the registered tools. Third, it is guided by a meta-prompt that specifies the objective of the current stage and the required format of the execution instruction. Accordingly, the input context $X_k$ at execution stage $k$ is represented as:
\begin{equation}
    X_k = \{P_0, H_k, \mathcal{F}(M), \Pi_k\}
\end{equation}
where $H_k$ denotes the accumulated execution history and intermediate results, $\mathcal{F}(M)$ denotes the functional descriptions and interface definitions of the tools registered in $M$, and $\Pi_k$ denotes the meta-prompt for the current stage.

Given $X_k$, the Task Engine determines the operation required at the current stage and converts it into an executable instruction. Depending on the execution state, this may involve invoking a prompt-analysis tool, configuring one or more enhancement tools, establishing dependencies among tool calls, or performing prompt fusion. When multiple tools are involved, the Task Engine also specifies their invocation order and data dependencies. The generated instruction $Y_k$ is formulated as:
\begin{equation}
    Y_k = E(X_k)
\end{equation}
where $Y_k$ specifies the tools to be invoked, their required arguments, and any dependencies on previously generated results. Therefore, the output of the Task Engine is not a visual prediction or the final answer $R'$, but an execution instruction that can be carried out by the registered tools.

Before execution, each generated instruction is checked by a format-validation function to ensure that its tool names, required arguments, and structure comply with the predefined interfaces. If $Y_k$ violates any constraint, an explicit error message $e_k$ is returned to the Task Engine, which regenerates the instruction for the current stage based on the original context and validation feedback:
\begin{equation}
    Y_k' = E(X_k, e_k)
\end{equation}
This validation--regeneration process is repeated until a valid instruction is obtained. It prevents malformed tool calls from being propagated to subsequent execution stages. For simplicity, the validated instruction is still denoted as $Y_k$ in the following formulation.

The validated instruction is then executed using the registered tools in the Tool Repository, producing the corresponding output $O_k$:
\begin{equation}
    O_k = \text{Exec}(Y_k, M)
\end{equation}
where $\text{Exec}(\cdot)$ denotes the instruction execution operation. The meaning of $O_k$ depends on the current stage. During Task Prompt Analysis, $O_k$ denotes the structured task-analysis results; during Tool Selection, it denotes the resulting tool-execution configuration; and during Enhanced Prompt Generation, the final output corresponds to the enhanced task prompt $P_e$. The detailed contents and structures of these stage-specific outputs are described in the following sections.

Each output is subsequently appended to the execution history:
\begin{equation}
    H_{k+1} = H_k \oplus O_k
\end{equation}
where $\oplus$ denotes the accumulation of newly generated results into the execution history. The updated context is then used by the Task Engine to determine the next operation, and this iterative process continues until prompt enhancement is completed.

\begin{table*}[t]
	\centering
	\caption{The detailed definition and usage of the tools in the Tool Repository.}
	\includegraphics[width=1.0\linewidth]{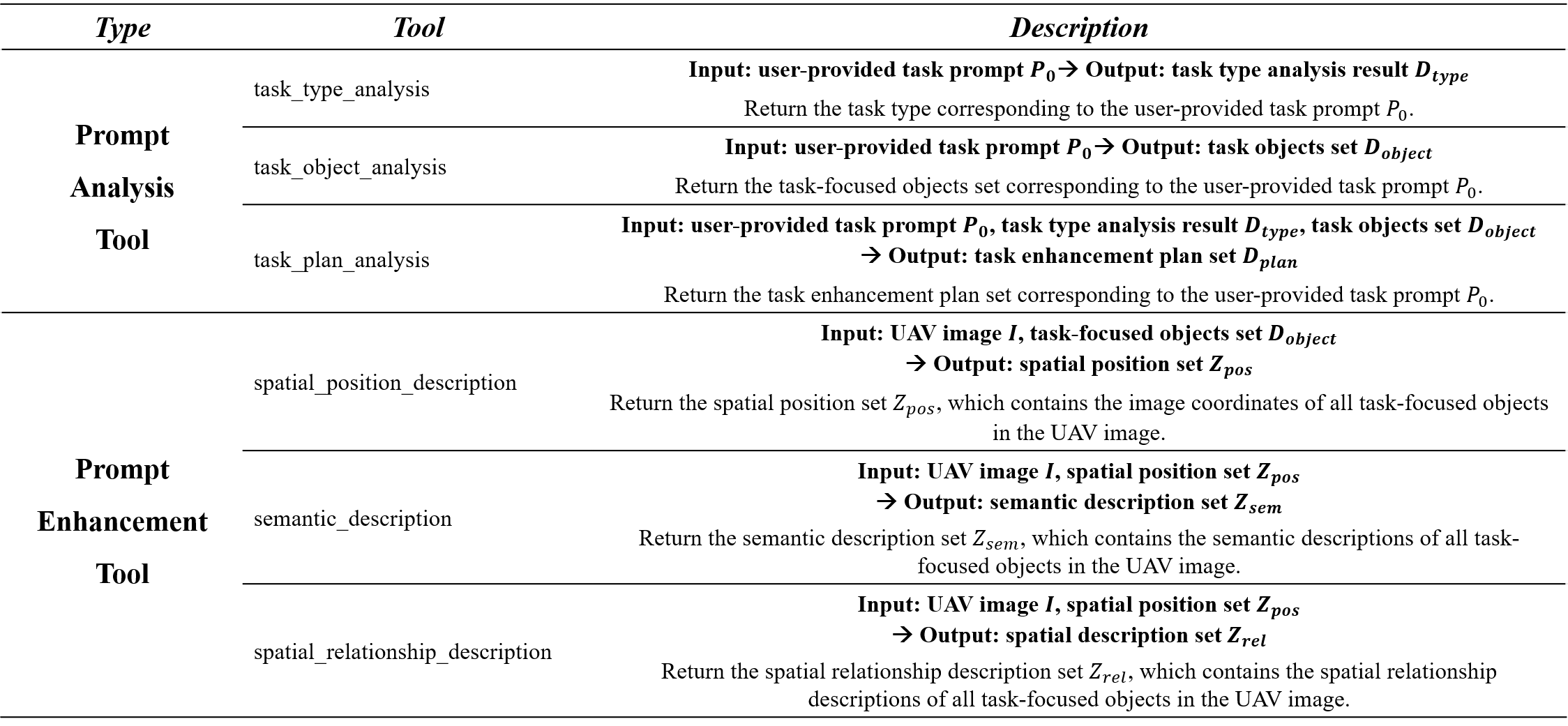}
	\label{tab:tool_definition}
\end{table*}

\subsubsection{Module 2. Tool Repository}\hfill\\
\indent The Tool Repository $M$ functions as the central hub for managing all specialized tools within the AerialVP framework. To support adaptive prompt enhancement for diverse UAV perception tasks, we construct a unified and modular Tool Repository that explicitly decouples prompt analysis from prompt enhancement. As illustrated in \textcolor{blue}{Fig. \ref{fig:tool_tree}}, the repository is organized into two functional categories: Prompt Analysis Tools and Prompt Enhancement Tools. Importantly, the Tool Repository is a modular tool-management framework constructed as part of AerialVP rather than a pre-existing toolkit. Most underlying tools are existing pretrained models or algorithms, together with customized functional modules, which are integrated into the repository through standardized interfaces. Accordingly, the contribution of the Tool Repository lies not in developing new underlying models, but in providing a unified mechanism for their registration, organization, invocation, and replacement. This modular design allows individual tools to be flexibly updated or extended without modifying the overall AerialVP workflow.

The Prompt Analysis Tools are designed to parse the original user-provided task prompt and extract the structured information required for downstream enhancement planning. In the current implementation, all three tools are instantiated using GPT-4o and accessed through cloud API calls. As summarized in \textcolor{blue}{Table \ref{tab:tool_definition}}, this category contains the following specialized tools:

\begin{itemize}
    \item \textbf{Task Type Analysis Tool (task\_type\_analysis)}: This tool identifies the perception task specified by the original prompt and returns the corresponding task type, such as Visual Grounding, Visual Reasoning, or Visual Question Answering.
    \item \textbf{Task Object Analysis Tool (task\_object\_analysis)}: This tool extracts the target and reference objects involved in the task and returns them as a structured object list, such as \textit{[“house”, “car”, “building”]}.
    \item \textbf{Task Enhancement Plan Analysis Tool (task\_plan\_\linebreak analysis)}: This tool determines which enhancement dimensions are required based on the current task and the original prompt, such as semantic description, spatial-position description, or spatial-relationship description, and returns the corresponding structured enhancement plan for subsequent Tool Selection.
\end{itemize}

The Prompt Enhancement Tools are designed to enhance the original task prompt by extracting task-relevant semantic and spatial information from UAV imagery. In contrast to the API-based Prompt Analysis Tools, the models used in this category are open-source and deployed locally, enabling UAV-image processing to remain within the local environment. The Prompt Enhancement Tools are organized into three functional categories:

\begin{itemize}
    \item \textbf{Spatial Position Description Tool (spatial\_position\_\linebreak description)}: This tool identifies task-relevant objects in the UAV image and provides their spatial locations using bounding-box coordinates. YOLO-World-X \cite{cite51} and Grounding-DINO \cite{cite52} are integrated as candidate object detectors.
    \item \textbf{Semantic Description Tool (semantic\_description)}: This tool generates detailed textual descriptions of task-relevant visual objects, including their appearance attributes and surrounding scene context. The repository provides DAM \cite{cite53}, LLaVA-1.5-7B \cite{cite54}, and LLaVA-OneVision-7B \cite{cite55} as candidate implementations.
    \item \textbf{Spatial Relationship Description Tool (spatial\_re-\linebreak lationship\_description)}: This tool extracts inter-object spatial relationships, including relative position, distance, and neighborhood relations. The repository supports SpaceOm \cite{cite56}, LLaVA-1.5-7B, and LLaVA-OneVision-7B as candidate implementations.
\end{itemize}

As detailed in \textcolor{blue}{Table \ref{tab:tool_definition}}, the inputs and outputs of each registered tool are explicitly defined. To support unified interaction among heterogeneous models and algorithms, the Tool Repository is implemented under the MCP Client-Server architecture. Each tool is encapsulated as an independent module with a standardized JSON Schema-based input/output interface, through which its functional description, required arguments, and output format are exposed to the Task Engine.

\begin{figure*}[ht]
	\centering
	\includegraphics[width=0.86\linewidth]{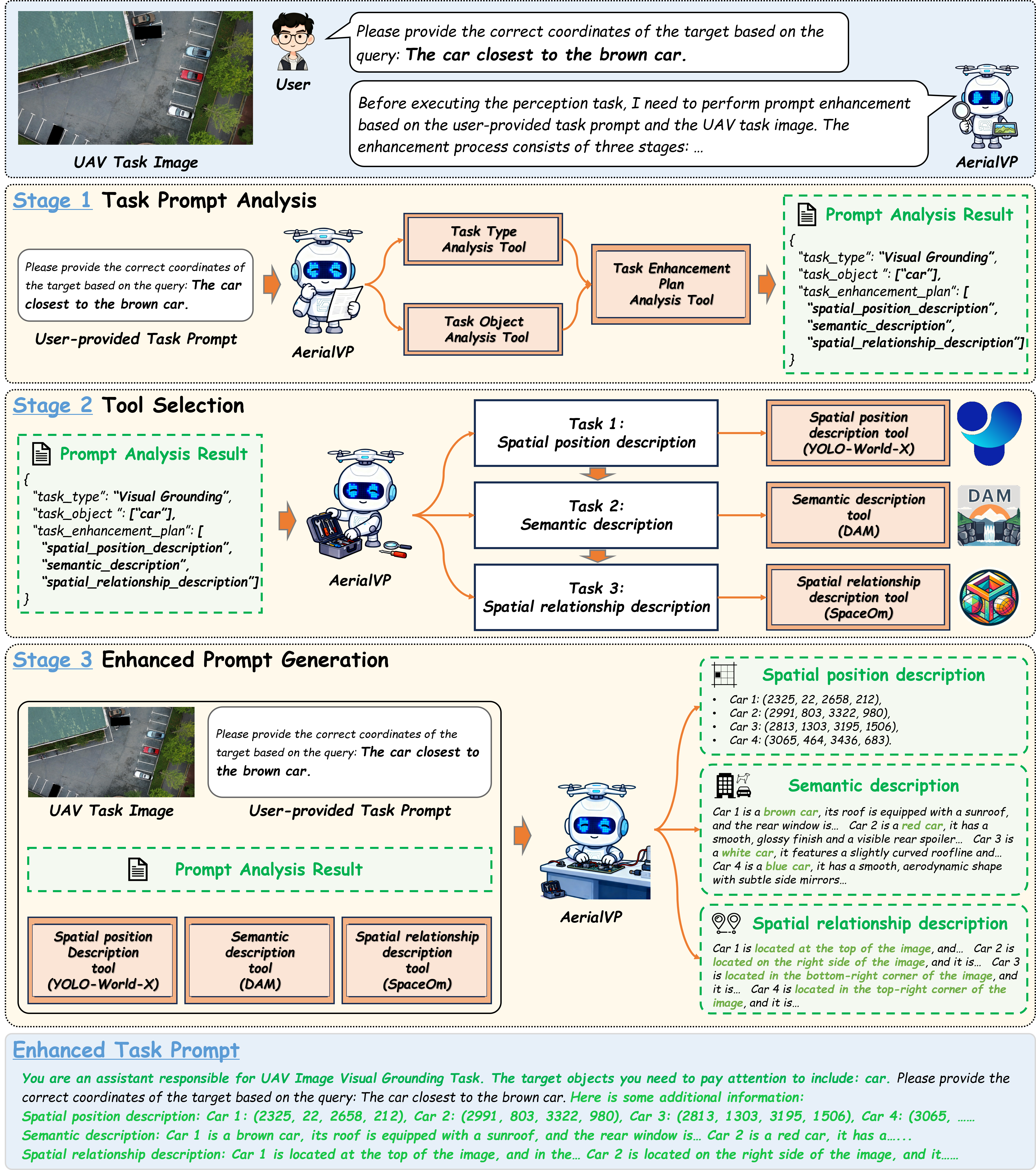}
	\vspace{-1mm}
	\caption{Workflow of the AerialVP agent for task prompt enhancement, consisting of three stages: Task Prompt Analysis, Tool Selection, and Enhanced Prompt Generation. The process is illustrated with the example of a vehicle localization task.}
	\vspace{-2mm} 
	\label{fig:workflow_example}
\end{figure*}

\subsection{Overview of the Task Prompt Enhancement Workflow}
The complete prompt-enhancement workflow is illustrated in \textcolor{blue}{Fig. \ref{fig:workflow_example}}. Given an original task prompt $P_0$ and a UAV image $I$, AerialVP progressively constructs the enhanced prompt $P_e$ through three stages: \textbf{Task Prompt Analysis}, \textbf{Tool Selection}, and \textbf{Enhanced Prompt Generation}.

In the Task Prompt Analysis stage, the Task Engine invokes three Prompt Analysis Tools to analyze the original task prompt. Specifically, it first identifies the task type and extracts the task-relevant objects, and then determines the enhancement dimensions required for the current task based on these analysis results. The outputs of these analyses are aggregated into a structured representation, which explicitly describes what the task is, which objects should be considered, and what additional information is required. This structured representation is then passed to the Tool Selection stage.

In the Tool Selection stage, the Task Engine maps the identified enhancement requirements to the corresponding functional tools registered in Tool Repository. Specifically, the Task Engine determines which enhancement functions should be activated, configures their required inputs, and organizes the execution dependencies among the selected tools. The resulting tool-execution configuration therefore specifies which tools should be invoked and how they should cooperate to obtain the information required by the current task.

In the Enhanced Prompt Generation stage, the Task Engine executes the selected tool workflow on the UAV image. Specifically, the Prompt Enhancement Tools extract task-relevant semantic, spatial-position, and spatial-relationship information according to the configuration generated in the preceding stage. Intermediate outputs can be propagated among tools when required, enabling different types of auxiliary information to complement one another. After tool execution is completed, the extracted information is aggregated and integrated with the original task prompt through a predefined prompt-fusion function, producing the enhanced prompt $P_e$.

This three-stage design establishes a clear division of responsibilities within AerialVP: Task Prompt Analysis determines what information is needed, Tool Selection determines how the required information should be obtained, and Enhanced Prompt Generation extracts and integrates that information into the final task prompt. The detailed implementation of the three stages is presented in \textcolor{blue}{Section \ref{stage:analysis} -- \ref{stage:gen}}.

\begin{figure*}[ht]
\centering
\includegraphics[width=0.9\linewidth]{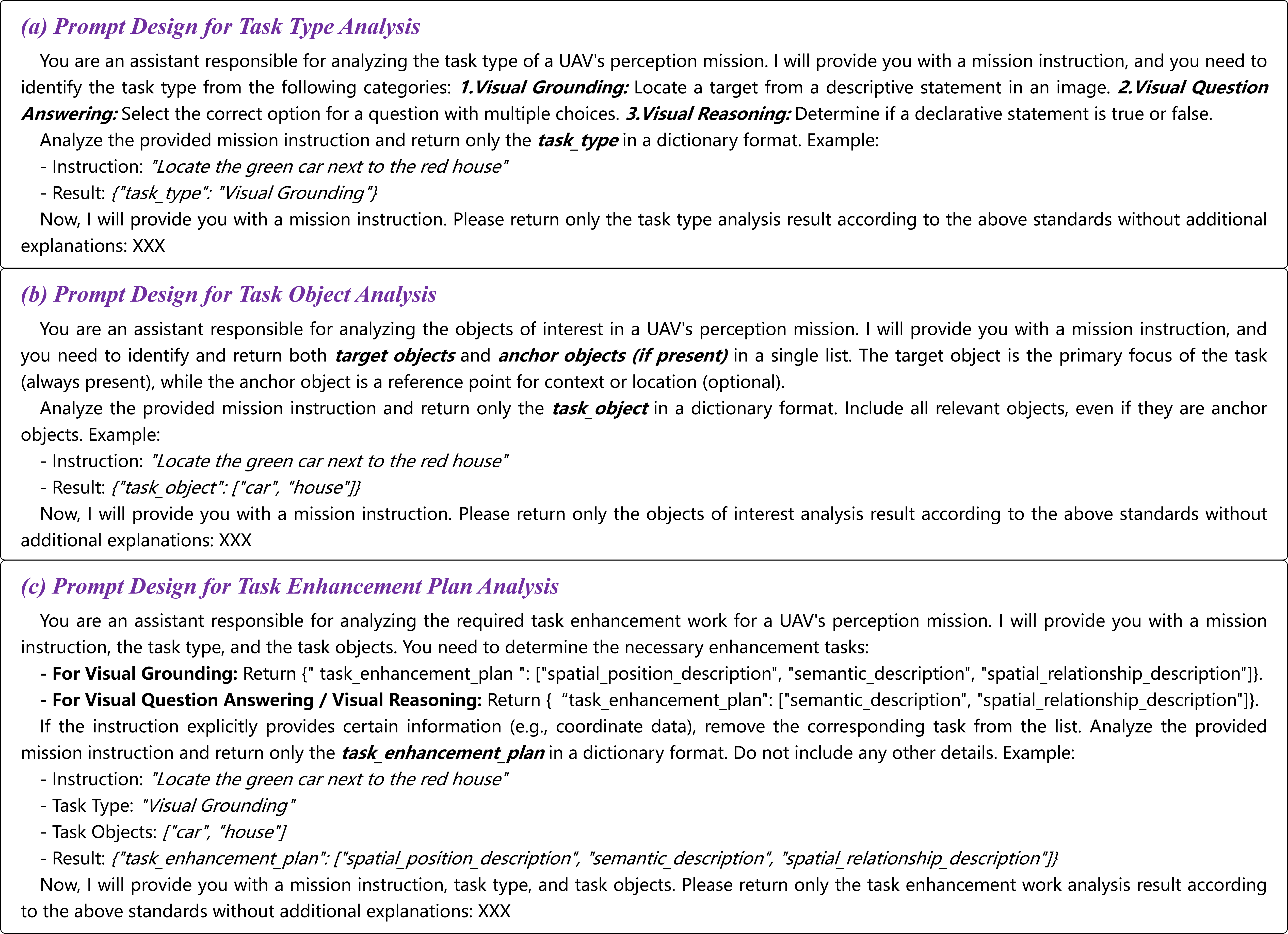}
\vspace{-1mm}
\caption{Tool-specific guidance prompts used by the Prompt Analysis Tools during Task Prompt Analysis stage. (a) Task Type Analysis guidance prompt $G_{\text{type}}$; (b) Task Object Analysis guidance prompt $G_{\text{object}}$; and (c) Task Enhancement Plan Analysis guidance prompt $G_{\text{plan}}$.}
\vspace{-2mm} 
\label{fig:guide_prompt}
\end{figure*}

\begin{figure*}[ht]
\centering
\includegraphics[width=0.9\linewidth]{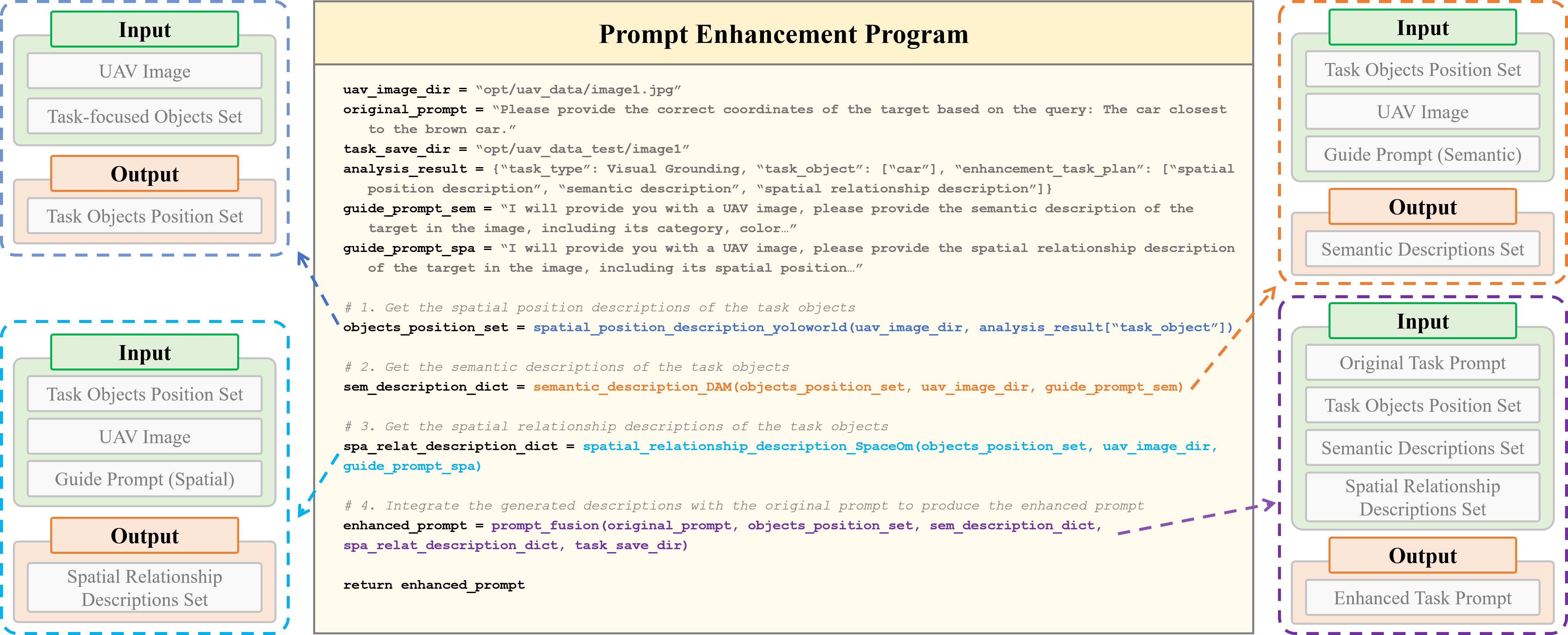}
\vspace{-1mm}
\caption{Example of the executable workflow generated for prompt enhancement in a vehicle-localization task. Intermediate outputs are propagated among the selected tools according to their information dependencies.}
\vspace{-2mm} 
\label{fig:program}
\end{figure*}

\subsection{Task Prompt Analysis}
\label{stage:analysis}
The objective of Task Prompt Analysis is to analyze the original task prompt and identify the information required for subsequent prompt enhancement. This task-specific analysis enables AerialVP to selectively incorporate relevant information instead of applying a fixed enhancement strategy. The input of this stage is the original task prompt $P_0$, while the output is a structured task representation containing the task type, task-relevant objects, and required enhancement dimensions.

As illustrated in \textcolor{blue}{Fig. \ref{fig:workflow_example}}, we decompose this stage into three analysis subtasks: Task Type Analysis, Task Object Analysis, and Task Enhancement Plan Analysis. The corresponding Prompt Analysis Tools are invoked using the tool-specific guidance prompts shown in \textcolor{blue}{Fig. \ref{fig:guide_prompt}}. The first two analyses extract the basic task information, while the third determines the enhancement requirements based on both the original prompt and the preceding analysis results.

Specifically, let $M_{\text{type}}, M_{\text{object}}, M_{\text{plan}} \in \mathcal{M}$ denote the Task Type Analysis Tool, Task Object Analysis Tool, and Task Enhancement Plan Analysis Tool, respectively. Given $P_0$ and their corresponding guidance prompts $G_{\text{type}}$, $G_{\text{object}}$, and $G_{\text{plan}}$, the analysis process is expressed as:
\begin{equation}
    D_{\text{type}} = M_{\text{type}}(P_0, G_{\text{type}})
\end{equation}
\begin{equation}
    D_{\text{object}} = M_{\text{object}}(P_0, G_{\text{object}})
\end{equation}
\begin{equation}
    D_{\text{plan}} = M_{\text{plan}}(P_0, D_{\text{type}}, D_{\text{object}}, G_{\text{plan}})
\end{equation}
where $D_{\text{type}}$ denotes the identified task type, such as Visual Grounding, Visual Reasoning, or Visual Question Answering; $D_{\text{object}}$ denotes the set of task-relevant objects; and $D_{\text{plan}}$ denotes the required enhancement dimensions, including semantic, spatial-position, and spatial-relationship information. These outputs are subsequently aggregated into a structured task representation $S$:
\begin{equation}
    S = \{D_{\text{type}}, D_{\text{object}}, D_{\text{plan}}\}
\end{equation}
For the vehicle-localization example illustrated in \textcolor{blue}{Fig. \ref{fig:workflow_example}}, the analysis produces \texttt{"Visual Grounding"} as the task type, [\texttt{"car"}] as the task-relevant object set, and [\texttt{"spatial\_position\_description"}, \texttt{"semantic\_description"}, \texttt{"spatial\_\linebreak relationship\_description"}] as the enhancement plan. The resulting representation $S$ then serves as the input to the subsequent Tool Selection stage.

\subsection{Tool Selection}
\label{stage:tool}
The objective of Tool Selection is to determine the functional tools required to satisfy the enhancement requirements identified in the preceding stage. The input of this stage is the structured task representation $S$, while the output is a tool-execution configuration specifying the selected tools, their invocation parameters, and their execution dependencies. Based on $S$, particularly the enhancement dimensions specified in $D_{\text{plan}}$, the Task Engine $E$ maps each information requirement to the corresponding functional tool registered in the Tool Repository $M$. This mapping relies on the standardized tool descriptions and interface definitions exposed by the MCP-based Tool Repository. The resulting tool-execution configuration is formulated as:
\begin{equation}
    C = E(S, \mathcal{F}(M), \Pi_{\text{sel}})
\end{equation}
where $\Pi_{\text{sel}}$ denotes the stage-specific meta-prompt for Tool Selection, and $C$ denotes the resulting configuration containing the selected tools, their invocation arguments, and execution dependencies.

For the vehicle-localization example, all three enhancement dimensions are required. Under the default configuration adopted in this work, these functions are instantiated by YOLO-World-X, DAM, and SpaceOm, respectively. The selected tools, along with their required inputs and execution dependencies, are organized into a structured configuration that serves as the input to the subsequent Enhanced Prompt Generation stage.

\subsection{Enhanced Prompt Generation}
\label{stage:gen}
The final stage of the workflow is Enhanced Prompt Generation. Its objective is to execute the selected Prompt Enhancement Tools and integrate the extracted task-relevant information with the original prompt $P_0$. To maintain semantic compatibility with the backend VLM, the extracted information is represented using common concept-level descriptions generated by general-purpose pretrained description models. The inputs of this stage are the tool-execution configuration generated in \textcolor{blue}{Section \ref{stage:tool}}, the original prompt $P_0$, and the UAV image $I$, while the output is the enhanced task prompt $P_e$. Specifically, this process consists of Tool Execution and Information Extraction, followed by a Prompt Fusion stage.

\subsubsection{Tool Execution and Information Extraction}\hfill\\
\indent Given the tool-execution configuration $C$, the Task Engine $E$ generates an executable workflow that specifies the selected tools, their invocation order, required arguments, and data dependencies. When multiple enhancement tools are involved, the output of a preceding tool can be directly propagated to subsequent tools according to the configured dependencies.

Let $M_{\text{pos}}$, $M_{\text{sem}}$, and $M_{\text{rel}}$ denote the selected Spatial Position Description Tool, Semantic Description Tool, and Spatial Relationship Description Tool, respectively. For the default workflow used in this work, when all three enhancement dimensions are required, the execution process is represented as:
\begin{equation}
    Z_{\text{pos}} = M_{\text{pos}}(I, D_{\text{object}})
\end{equation}
\begin{equation}
    Z_{\text{sem}} = M_{\text{sem}}(I, Z_{\text{pos}})
\end{equation}
\begin{equation}
    Z_{\text{rel}} = M_{\text{rel}}(I, Z_{\text{pos}})
\end{equation}
where $Z_{\text{pos}}$, $Z_{\text{sem}}$, and $Z_{\text{rel}}$ denote the extracted spatial-position, semantic, and spatial-relationship information, respectively.

As illustrated in \textcolor{blue}{Fig. \ref{fig:program}}, for the vehicle-localization example, the workflow first invokes \texttt{spatial\_position\_\linebreak description\_yoloworld} to obtain candidate target locations. The resulting coordinates are passed to \texttt{semantic\_description\_DAM} for target-specific semantic description and to \texttt{spatial\_relationship\_\linebreak description\_SpaceOm} for inferring spatial relationships between the target and surrounding objects. In this way, intermediate results are propagated through the selected tool-execution workflow according to their dependencies without manual intervention. After all required tools have completed execution, their outputs are collected and passed to the prompt-fusion function.

\subsubsection{Prompt Fusion}\hfill\\
\indent Prompt Fusion integrates the extracted auxiliary information with the original task prompt while preserving its original task objective. Let $Z = \{Z_{\text{pos}}, Z_{\text{sem}}, Z_{\text{rel}}\}$ denote the available outputs of the selected Prompt Enhancement Tools. The enhanced prompt is generated as:
\begin{equation}
    P_e = U(P_0, D_{\text{type}}, D_{\text{object}}, Z)
\end{equation}
where $U(\cdot)$ denotes the predefined prompt-fusion function. In implementation, $P_0$ is retained as the core instruction, while the identified task type, task objects, and selected enhancement information are incorporated using the following template:

\textit{“You are an assistant responsible for UAV Image [\(task\_type\)] Task. The target objects you need to pay attention to include: [\(task\_object\)]. [\(Original\_task\_instruction\)]. Here is some additional information: \newline [\(Enhancement\_information\)]”}

Only the enhancement dimensions identified during Task Prompt Analysis are included in the final prompt. Therefore, the content of $P_e$ is adapted to the information requirements of the current task rather than following a fixed combination of all enhancement types. As shown at the bottom of \textcolor{blue}{Fig. \ref{fig:workflow_example}}, the vehicle-localization example retains the original task instruction and supplements it with the semantic, positional, and relational information generated by the selected tools. This predefined fusion mechanism makes explicit how the outputs of individual tools are incorporated into the final enhanced prompt.

Once $P_e$ is constructed, the prompt-enhancement process is complete. The enhanced prompt $P_e$ and the original UAV image $I$ are then provided to the backend VLM to obtain the final prediction $R'$ according to \textcolor{blue}{Fig. \ref{eq:enhanced}}.


\begin{figure*}[ht]
\centering
\includegraphics[width=0.9\linewidth]{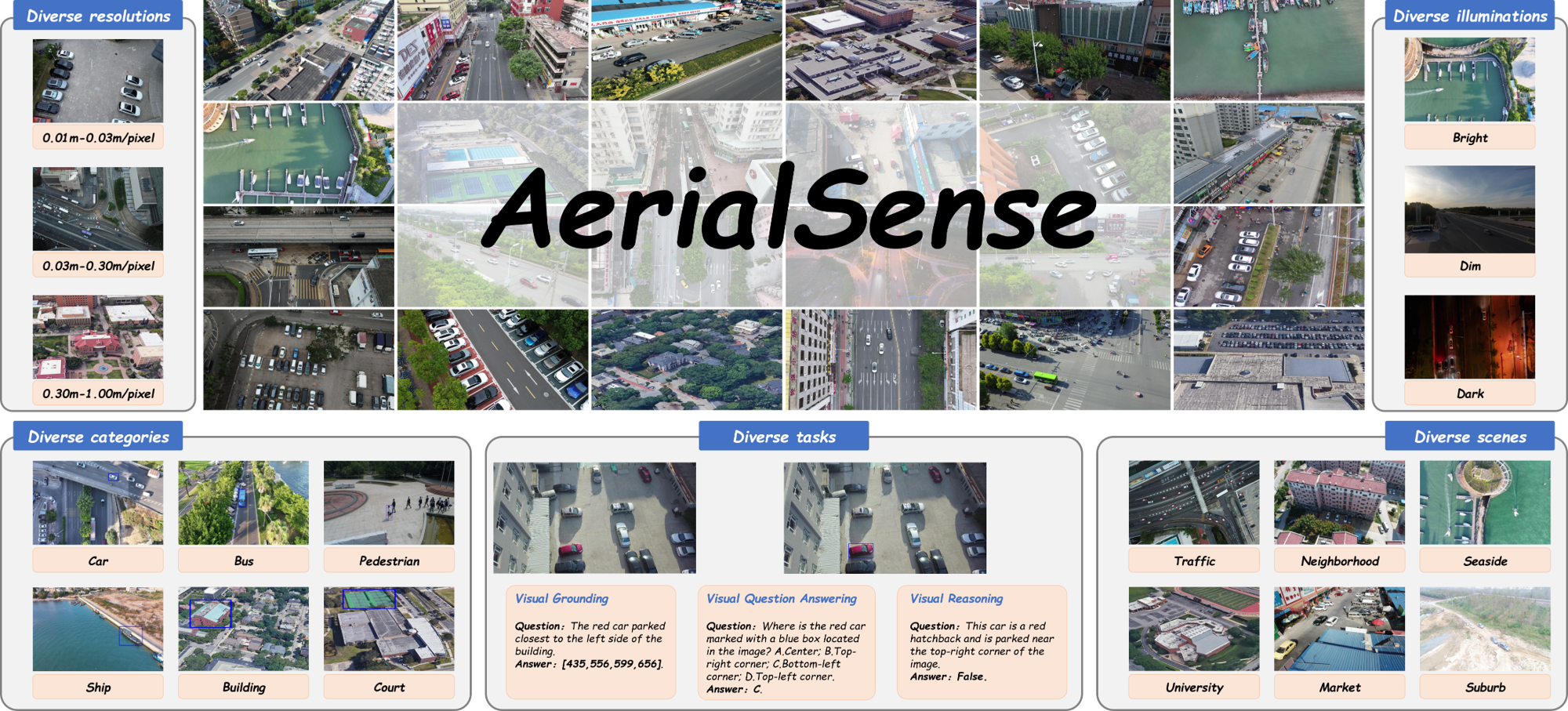}
\vspace{-1mm}
\caption{Overview of the AerialSense benchmark. AerialSense integrates multiple publicly available UAV datasets through a unified annotation and instruction-transformation pipeline. It supports three representative tasks: Aerial Visual Grounding, Aerial Visual Reasoning, and Aerial Visual Question Answering.}
\vspace{-2mm} 
\label{fig:dataset}
\end{figure*}

\begingroup
\newcolumntype{M}[1]{>{\raggedright\arraybackslash}m{#1}}
\newcolumntype{C}[1]{>{\centering\arraybackslash}m{#1}}
\renewcommand{\arraystretch}{1.4} 

\begin{table*}[htbp]
\centering
\caption{Comparison of AerialSense with existing UAV image perception datasets}
\label{tab:dataset_comparison}
\begin{tabular}{M{3.0cm} C{2.2cm} C{2.5cm} C{2.5cm} C{2.6cm} C{2.6cm}}
\toprule
\textbf{Dataset Name} & 
\textbf{Task Types} & 
\textbf{Number of UAV Images} & 
\textbf{Number of Task Samples} & 
\textbf{Number of Target Categories} & 
\textbf{Average Instruction Length} \\
\midrule
HurMic-VQA \cite{cite64} & VQA & 3,197 & 9,591 & 4 & 7.11 \\
FloodNet-VQA \cite{cite65} & VQA & 2,188 & 7,355 & 9 & 7.89 \\
FloodNet-VQA V2.0 \cite{cite66} & VQA & 2,348 & 10,480 & 9 & 7.93 \\
refGeo \cite{cite59} & VG & 623 & 26,482 & 1 & 5.58 \\
AerialVG \cite{cite57} & VG & 4,211 & 4,723 & 26 & 19.75 \\
AerialSense & \textbf{VG, VQA, VR} & \textbf{7,119} & \textbf{53,374} & \textbf{40} & \textbf{20.76} \\
\bottomrule
\end{tabular}
\end{table*}
\endgroup

\begin{figure*}[ht]
\centering
\includegraphics[width=0.95\linewidth]{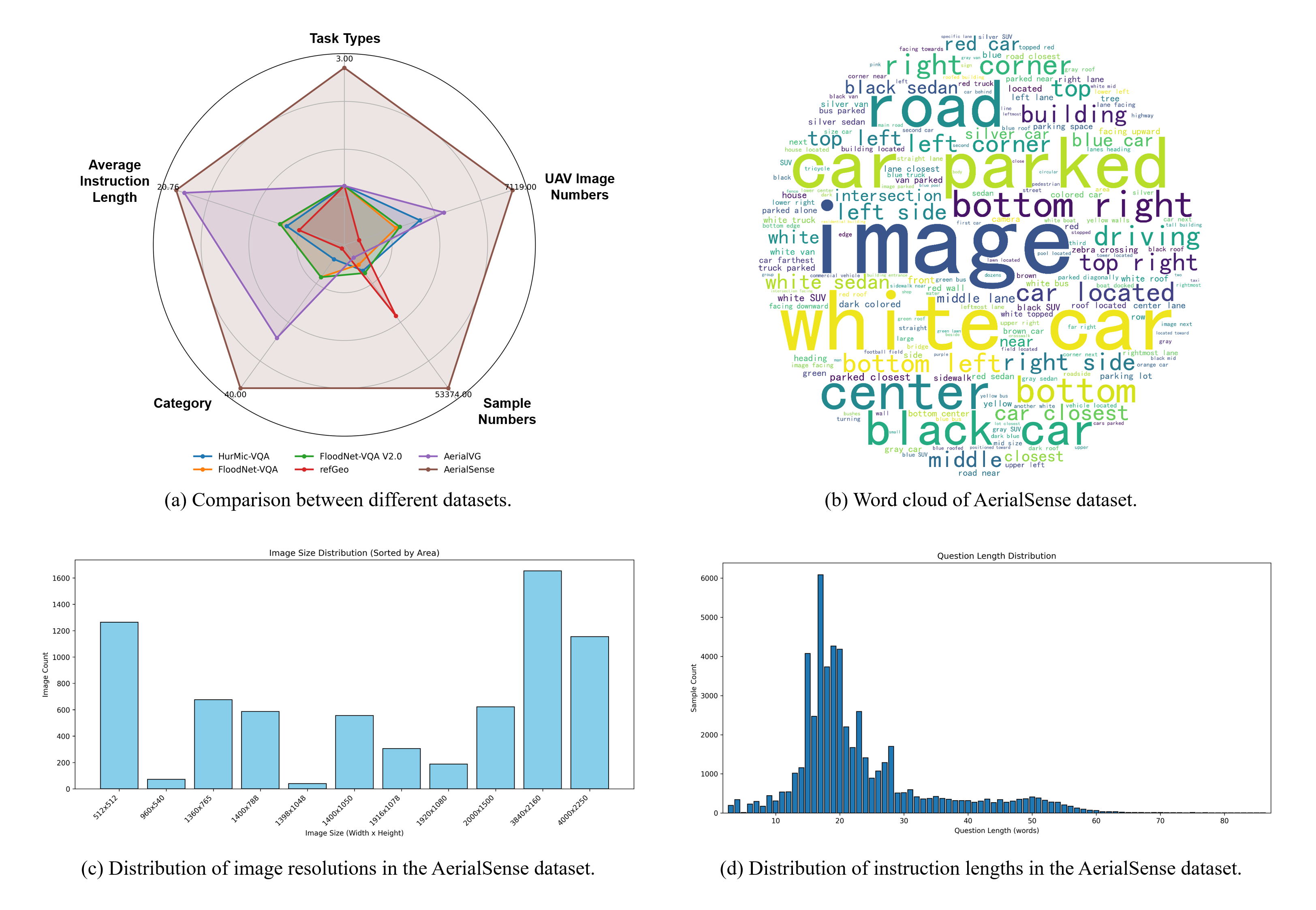}
\vspace{-1mm}
\caption{Statistical overview of the AerialSense benchmark, including (a) comparison with existing UAV perception datasets, (b) instruction word cloud, (c) distribution of image resolutions, and (d) distribution of instruction lengths.}
\vspace{-2mm} 
\label{fig:dataset_info}
\end{figure*}

\section{AerialSense: A Comprehensive Benchmark for UAV Image Perception}
Existing UAV perception datasets often suffer from limited task coverage and scene diversity, typically focusing on a single task and a narrow range of targets and imaging conditions. This limits comprehensive evaluation of VLMs across diverse UAV scenarios. To address these issues, we construct AerialSense, a unified multi-task benchmark for UAV image perception.

As illustrated in \textcolor{blue}{Fig. \ref{fig:dataset}}, AerialSense supports three representative tasks: \textbf{Aerial Visual Grounding (VG)}, \textbf{Aerial Visual Question Answering (VQA)}, and \textbf{Aerial Visual Reasoning (VR)}. Specifically, AerialSense integrates multiple publicly available UAV datasets and reorganizes them through a unified annotation and instruction-transformation pipeline. The three tasks are built on a shared object-centric annotation foundation, ensuring consistent target definitions across different task formulations. In total, AerialSense contains 7,119 UAV images and 53,374 task samples, providing a standardized platform for evaluating VLM localization, understanding, and reasoning capabilities under diverse UAV scenarios.

\subsection{Dataset Sources}
To ensure broad scene and target coverage, AerialSense integrates representative UAV and aerial imagery from five publicly available datasets with complementary characteristics:

\begin{itemize}
    \item \textbf{AerialVG} \cite{cite57}: AerialVG is constructed from high-resolution VisDrone \cite{cite58} images and provides fine-grained referring expressions containing semantic attributes, spatial orientations, and relational cues. We retain its language and localization annotations as an important source for constructing complex grounding samples.
    \item \textbf{refGeo} \cite{cite59}: refGeo integrates established remote-sensing Visual Grounding datasets, including GeoChat \cite{cite60} and VRSBench \cite{cite61}. We utilize its referring expressions and bounding-box annotations to enrich the localization diversity of AerialSense.
    \item \textbf{CODrone} \cite{cite62}: CODrone is included to mitigate the vehicle-centric category bias of AerialVG and refGeo. Its 12 object categories, including pedestrians and traffic-related objects, broaden the target coverage of the benchmark.
    \item \textbf{VisDrone} \cite{cite58}: We sample from the image-based object-detection subset of VisDrone to increase scene and object diversity. Since AerialVG is also derived from VisDrone, overlapping samples are removed through cross-filtering to avoid duplication and cross-source overlap.
    \item \textbf{University-1652} \cite{cite63}: University-1652 provides multi-view UAV imagery with diverse architectural structures. We selectively sample approximately 10\% of its UAV test images to increase building diversity, viewpoint variation, and image-resolution coverage.
\end{itemize}

\subsection{Benchmark Construction Pipeline}
To transform the heterogeneous source datasets into a unified multi-task benchmark, we process the collected aerial images through a structured four-stage construction pipeline. The overall process consists of Image Filtering and Selection, Visual Grounding Construction, Instruction Transformation, and Quality Verification. To ensure consistency across different tasks, AerialSense first establishes standardized object-centric grounding annotations and then derives VR and VQA samples from the same annotation foundation. GPT-4o is used only as an auxiliary tool for grammatical checking and expression refinement, while the dataset reconstruction and verification procedures are manually conducted.

\textbf{Stage 1: Image Filtering and Selection}\hfill\\
\indent To construct a high-quality and diverse UAV image collection, we first remove low-quality images with severe blur, insufficient target information, or limited scene content, reducing the impact of unreliable visual samples on subsequent task construction. On this basis, we further select and organize samples across multiple dimensions, including scene type (e.g., urban roads, seaside areas, suburbs, and markets), lighting conditions (e.g., bright, dim, and dark), object categories (e.g., vehicles, pedestrians, courts, and buildings), and image resolutions. This process produces a representative UAV image collection with diverse environments for subsequent task construction.

\textbf{Stage 2: Visual Grounding Construction}\hfill\\
\indent In this stage, the filtered images are converted into standardized VG samples, which form the common object-centric annotation foundation of AerialSense. For datasets with existing grounding annotations, including AerialVG and refGeo, the original referring expressions and bounding-box annotations are retained and standardized into a unified format. For datasets without grounding annotations, we construct grounding samples according to the availability of localization information. When object detection annotations are available, such as in CODrone and VisDrone, a subset of detected objects is selected as grounding targets, and annotators manually generate corresponding referring expressions. For datasets without localization annotations, such as University-1652, annotators manually select discriminative targets from the images, annotate their bounding boxes, and generate corresponding referring expressions. All referring expressions describe targets using object categories, appearance attributes, and spatial constraints, enabling the target objects to be uniquely identified in UAV images. The resulting annotations establish consistent correspondences among UAV images, target objects, referring expressions, and spatial locations, and provide the basis for constructing the other two task types.

\textbf{Stage 3: Instruction Transformation}\hfill\\
\indent Rather than independently annotating VR and VQA samples, AerialSense derives both tasks from the standardized grounding annotations generated in Stage 2. For each grounding sample, the annotated bounding box is overlaid on the original UAV image to highlight the target object, and the resulting marked image is used as the input image for both the VR and VQA tasks. The corresponding verification statements and questions are then constructed around the highlighted target. This strategy ensures that VG, VR, and VQA share consistent target objects and underlying semantic-spatial information while presenting them through different task formulations.

For VR tasks, the grounding descriptions are transformed into binary verification statements. Positive samples preserve the semantic and spatial information of the original annotation and are labeled \texttt{True}, whereas negative samples are generated by modifying semantic attributes or spatial relationships and are labeled \texttt{False}. For example, the grounding description \textit{“the red car closest to the building on the left”} can be retained as a positive statement, while changing its appearance attribute (\textit{“red”} to \textit{“pink”}) or spatial relation (\textit{“left”} to \textit{“right”}) produces a corresponding negative statement. These negative samples are retained only when the modified statements are inconsistent with the underlying annotations.

For VQA tasks, the grounding descriptions are reformulated into multiple-choice questions. The correct answer is derived from the annotated target information, while distractor options are selected from surrounding objects in the same image to maintain visual and semantic relevance. Semantic and spatial attributes are further recombined where appropriate to generate additional instruction variants and increase task diversity.

Through this transformation strategy, the three tasks share the same underlying object annotations while evaluating different aspects of localization, visual understanding, and semantic-spatial reasoning.

\textbf{Stage 4: Quality Verification}\hfill\\
\indent Finally, all constructed samples undergo manual quality verification. Grounding samples are retained only when independent annotators can consistently identify the same target from the referring expression. VR and VQA samples are checked for semantic correctness, logical consistency, and consistency with the underlying object annotations. Samples that do not satisfy these criteria are revised or discarded, ensuring reliable image-text correspondence across the three task types.

\subsection{Benchmark Statistics}
\textcolor{blue}{Table \ref{tab:dataset_comparison}} compares AerialSense with existing UAV image perception datasets in terms of task coverage, data scale, target-category diversity, and instruction length. \textcolor{blue}{Fig. \ref{fig:dataset_info}} further presents the overall statistics of AerialSense, including dataset comparison, instruction word distribution, image resolutions, and instruction lengths.

\begin{itemize}
    \item \textbf{Comprehensive Task Coverage:} As shown in \textcolor{blue}{Fig. \ref{fig:dataset_info} (a)}, to the best of our knowledge, AerialSense is among the first UAV-oriented benchmarks to jointly support VG, VQA, and VR within a unified benchmark. Existing datasets such as HurMic-VQA \cite{cite64}, FloodNet-VQA \cite{cite65}, and FloodNet-VQA V2.0 \cite{cite66} mainly focus on VQA, whereas refGeo and AerialVG focus on VG. By integrating localization, question answering, and reasoning tasks, AerialSense enables a more comprehensive evaluation of VLM perception capabilities in UAV scenarios.
    
    \item \textbf{Benchmark Scale and Object Diversity:} AerialSense contains 7,119 UAV images and 53,374 task samples, providing broader evaluation coverage than the compared UAV perception benchmarks. Beyond sample quantity, AerialSense also achieves broader object diversity with 40 object categories, as illustrated in \textcolor{blue}{Fig. \ref{fig:dataset_info} (a)}. The benchmark covers a wide range of UAV-relevant targets, including ground vehicles, pedestrians, buildings, ships, and land objects, reducing category concentration toward a limited set of target types. This diverse target distribution provides a broader evaluation setting for assessing the generalization ability of VLMs across different UAV perception scenarios.
    
    \item \textbf{Image Resolution Diversity:} As shown in \textcolor{blue}{Fig. \ref{fig:dataset_info} (c)}, AerialSense covers a wide range of image resolutions from $512 \times 512$ to $4000 \times 2250$, reflecting the diverse imaging configurations encountered in UAV imagery. Such resolution variation also contributes to differences in the apparent scale of observed targets, increasing the diversity of target representations across images.
    
    \item \textbf{Instruction Characteristics:} AerialSense contains task instructions with substantial semantic and spatial content. The average instruction length is 20.76 words, compared with 19.75 words for AerialVG, 7.11 words for HurMic-VQA, and 5.58 words for refGeo. As shown in \textcolor{blue}{Fig. \ref{fig:dataset_info} (d)}, most instructions contain approximately 15--25 words, while some extend beyond 80 words. The word cloud in \textcolor{blue}{Fig. \ref{fig:dataset_info} (b)} further shows frequent references to object categories, appearance attributes, and spatial positions, reflecting the frequent use of semantic attributes and spatial descriptions in the task instructions.
\end{itemize}


\begingroup
\newcolumntype{M}[1]{>{\raggedright\arraybackslash}m{#1}}
\newcolumntype{C}[1]{>{\centering\arraybackslash}m{#1}}
\renewcommand{\arraystretch}{1.4} 

\begin{table*}[ht]
\centering
\caption{Experimental results on Aerial Visual Grounding (VG) tasks}
\label{tab:VG_acc}
\begin{tabular}{M{3.0cm} C{2.5cm} C{2.5cm} C{2.5cm} C{2.0cm} C{2.0cm}}
\toprule
Task Model & Semantic Enhancement & Spatial Position Enhancement & Spatial Relationship Enhancement & Acc & Mean IoU \\
\midrule
\cellcolor{lightgray}\textit{Proprietary Models} & \cellcolor{lightgray} & \cellcolor{lightgray} & \cellcolor{lightgray} & \cellcolor{lightgray} & \cellcolor{lightgray} \\
\hline
\multirow{2}{*}{GPT-4o} &  $\times$ & $\times$ & $\times$ & 2.04\% & 3.69\% \\
& $\checkmark$ & $\checkmark$ & $\checkmark$ & 45.05\% & 41.26\% \\ \hline
\multirow{2}{*}{GPT-4.1} & $\times$ & $\times$ & $\times$ & 3.60\% & 11.91\% \\
& $\checkmark$ & $\checkmark$ & $\checkmark$ & 49.28\% & 44.87\% \\ \hline
\multirow{2}{*}{Gemini-2.0-Flash} & $\times$ & $\times$ & $\times$ & 10.17\% & 14.99\% \\
& $\checkmark$ & $\checkmark$ & $\checkmark$ & 51.09\% & 46.22\% \\ \hline
\multirow{2}{*}{Claude-3.7-Sonnet} & $\times$ & $\times$ & $\times$ & 8.80\% & 12.94\% \\
& $\checkmark$ & $\checkmark$ & $\checkmark$ & 44.93\% & 41.08\% \\ \hline
\multirow{2}{*}{Qwen2.5-Max} & $\times$ & $\times$ & $\times$ & 1.22\% & 7.17\% \\
& $\checkmark$ & $\checkmark$ & $\checkmark$ & 43.87\% & 39.87\% \\
\midrule
\cellcolor{lightgray}\textit{Open-Source Models} & \cellcolor{lightgray} & \cellcolor{lightgray} & \cellcolor{lightgray} & \cellcolor{lightgray} & \cellcolor{lightgray} \\
\hline
\multirow{2}{*}{InternVL3-8B} & $\times$ & $\times$ & $\times$ & 2.40\% & 7.38\% \\
& $\checkmark$ & $\checkmark$ & $\checkmark$ & 14.21\% & 14.50\% \\ \hline
\multirow{2}{*}{InternVL3-14B} & $\times$ & $\times$ & $\times$ & 11.62\% & 18.36\% \\
& $\checkmark$ & $\checkmark$ & $\checkmark$ & 26.59\% & 24.48\% \\ \hline
\multirow{2}{*}{SAIL-VL-8B} & $\times$ & $\times$ & $\times$ & 6.15\% & 10.93\% \\
& $\checkmark$ & $\checkmark$ & $\checkmark$ & 34.23\% & 31.82\% \\ \hline
\multirow{2}{*}{Qwen2-VL-7B} & $\times$ & $\times$ & $\times$ & 17.94\% & 19.43\% \\
& $\checkmark$ & $\checkmark$ & $\checkmark$ & 37.94\% & 34.79\% \\ \hline
\multirow{2}{*}{LLaVA-OneVision-7B} & $\times$ & $\times$ & $\times$ & 8.63\% & 14.26\% \\
& $\checkmark$ & $\checkmark$ & $\checkmark$ & 38.65\% & 35.69\% \\ \hline
\multirow{2}{*}{LLaVA-1.5-7B} & $\times$ & $\times$ & $\times$ & 7.60\% & 13.57\% \\
& $\checkmark$ & $\checkmark$ & $\checkmark$ & 22.88\% & 21.61\% \\
\bottomrule
\end{tabular}
\end{table*}

\begin{table*}[ht]
\centering
\caption{Experimental results on Aerial Visual Reasoning (VR) and Aerial Visual Question Answering (VQA) tasks}
\label{tab:VRVQA_acc}
\begin{tabular}{M{2.8cm} C{3.0cm} C{4.4cm} C{2.0cm} C{2.0cm}}
\toprule
Task Model & Semantic Enhancement & Spatial Relationship Enhancement & Acc(VR) & Acc(VQA) \\
\midrule
\cellcolor{lightgray}\textit{Proprietary Models} & \cellcolor{lightgray} & \cellcolor{lightgray} & \cellcolor{lightgray} & \cellcolor{lightgray} \\
\hline
\multirow{2}{*}{GPT-4o} & $\times$ & $\times$ & 50.72\% & 60.03\% \\
& $\checkmark$ & $\checkmark$ & 80.71\% & 80.53\% \\ \hline
\multirow{2}{*}{GPT-4.1} & $\times$ & $\times$ & 70.40\% & 82.94\% \\
& $\checkmark$ & $\checkmark$ & 90.32\% & 96.73\% \\ \hline
\multirow{2}{*}{Gemini-2.0-Flash} & $\times$ & $\times$ & 75.65\% & 73.82\% \\
& $\checkmark$ & $\checkmark$ & 90.40\% & 95.35\% \\ \hline
\multirow{2}{*}{Claude-3.7-Sonnet} & $\times$ & $\times$ & 62.78\% & 71.70\% \\
& $\checkmark$ & $\checkmark$ & 85.48\% & 90.38\% \\ \hline
\multirow{2}{*}{Qwen2.5-Max} & $\times$ & $\times$ & 77.03\% & 65.95\% \\
& $\checkmark$ & $\checkmark$ & 90.48\% & 77.54\% \\
\midrule
\cellcolor{lightgray}\textit{Open-Source Models} & \cellcolor{lightgray} & \cellcolor{lightgray} & \cellcolor{lightgray} & \cellcolor{lightgray} \\
\hline
\multirow{2}{*}{InternVL3-8B} & $\times$ & $\times$ & 56.60\% & 74.91\% \\
& $\checkmark$ & $\checkmark$ & 78.85\% & 95.40\% \\ \hline
\multirow{2}{*}{InternVL3-14B} & $\times$ & $\times$ & 63.22\% & 74.93\% \\
& $\checkmark$ & $\checkmark$ & 80.30\% & 96.66\% \\ \hline
\multirow{2}{*}{SAIL-VL-8B} & $\times$ & $\times$ & 46.64\% & 25.84\% \\
& $\checkmark$ & $\checkmark$ & 73.84\% & 82.76\% \\ \hline
\multirow{2}{*}{Qwen2-VL-7B} & $\times$ & $\times$ & 56.40\% & 69.76\% \\
& $\checkmark$ & $\checkmark$ & 85.60\% & 93.74\% \\ \hline
\multirow{2}{*}{LLaVA-OneVision-7B} & $\times$ & $\times$ & 69.48\% & 73.37\% \\
& $\checkmark$ & $\checkmark$ & 83.91\% & 94.48\% \\ \hline
\multirow{2}{*}{LLaVA-1.5-7B} & $\times$ & $\times$ & 53.89\% & 51.64\% \\
& $\checkmark$ & $\checkmark$ & 90.44\% & 90.06\% \\
\bottomrule
\end{tabular}
\end{table*}
\endgroup

\begin{figure*}[ht]
\centering
\includegraphics[width=0.9\linewidth]{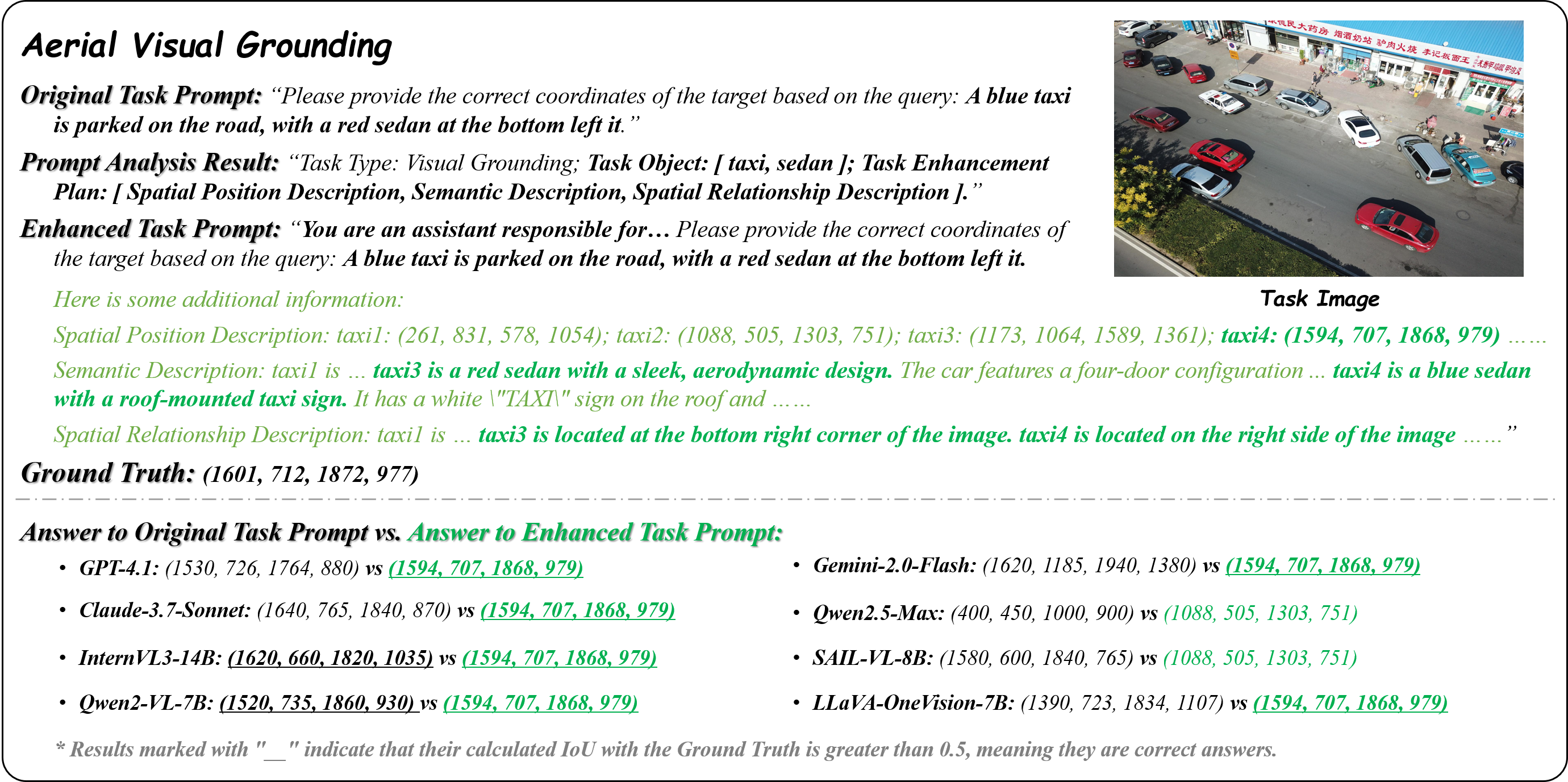}
\vspace{-1mm}
\caption{Illustration of the Aerial Visual Grounding task guided by enhanced prompts. Experimental results demonstrate that the enhanced prompts enable most tested models to accurately locate the task targets.}
\vspace{-2mm} 
\label{fig:VG_show}
\end{figure*}

\begin{figure*}[ht]
\centering
\includegraphics[width=0.9\linewidth]{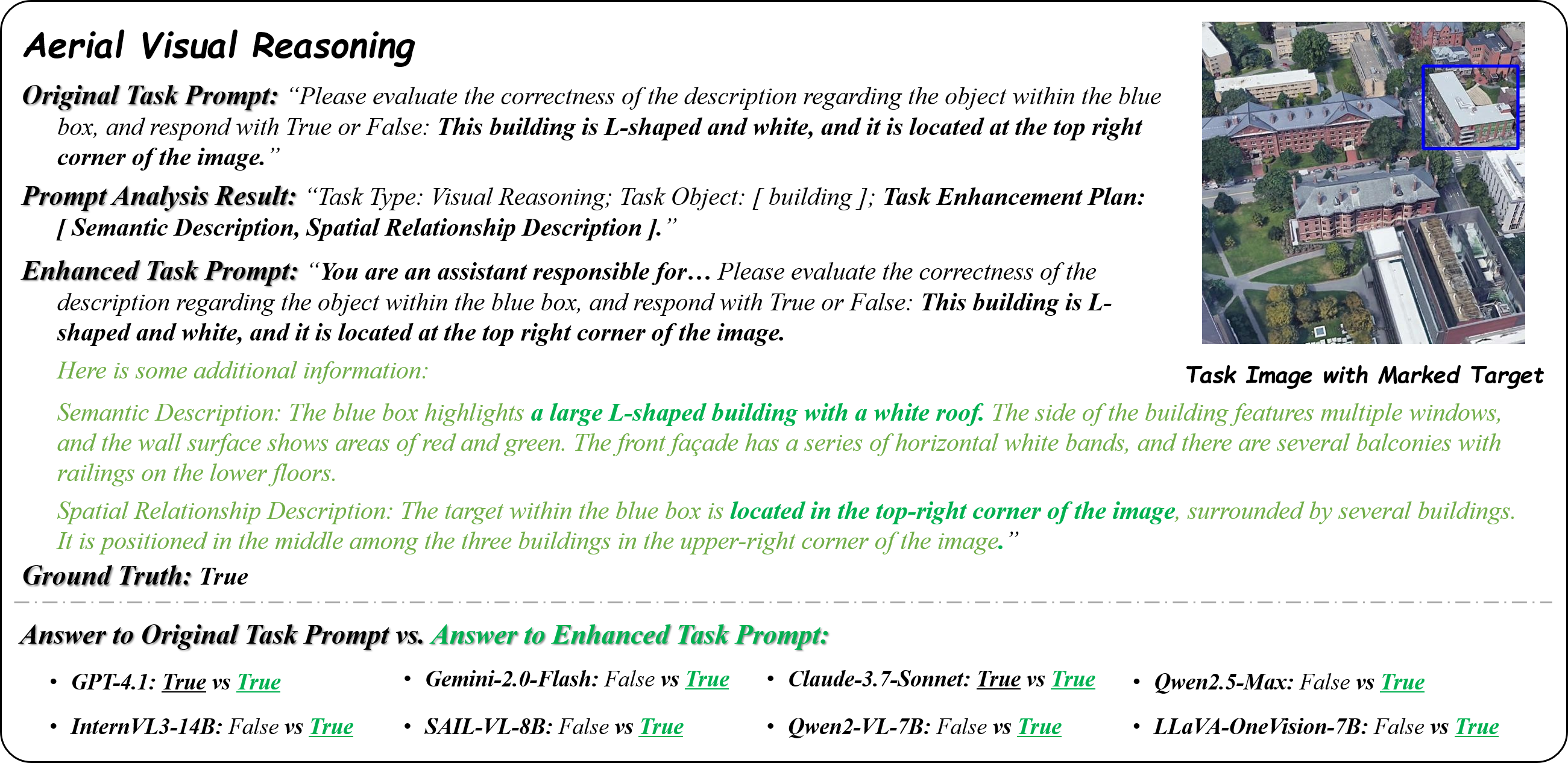}
\vspace{-1mm}
\caption{Illustration of the Aerial Visual Reasoning task guided by prompt enhancement. The results indicate that the enhanced prompts help most tested models achieve more accurate perception of visual information.}
\vspace{-2mm} 
\label{fig:VR_show}
\end{figure*}

\begin{figure*}[ht]
\centering
\includegraphics[width=0.9\linewidth]{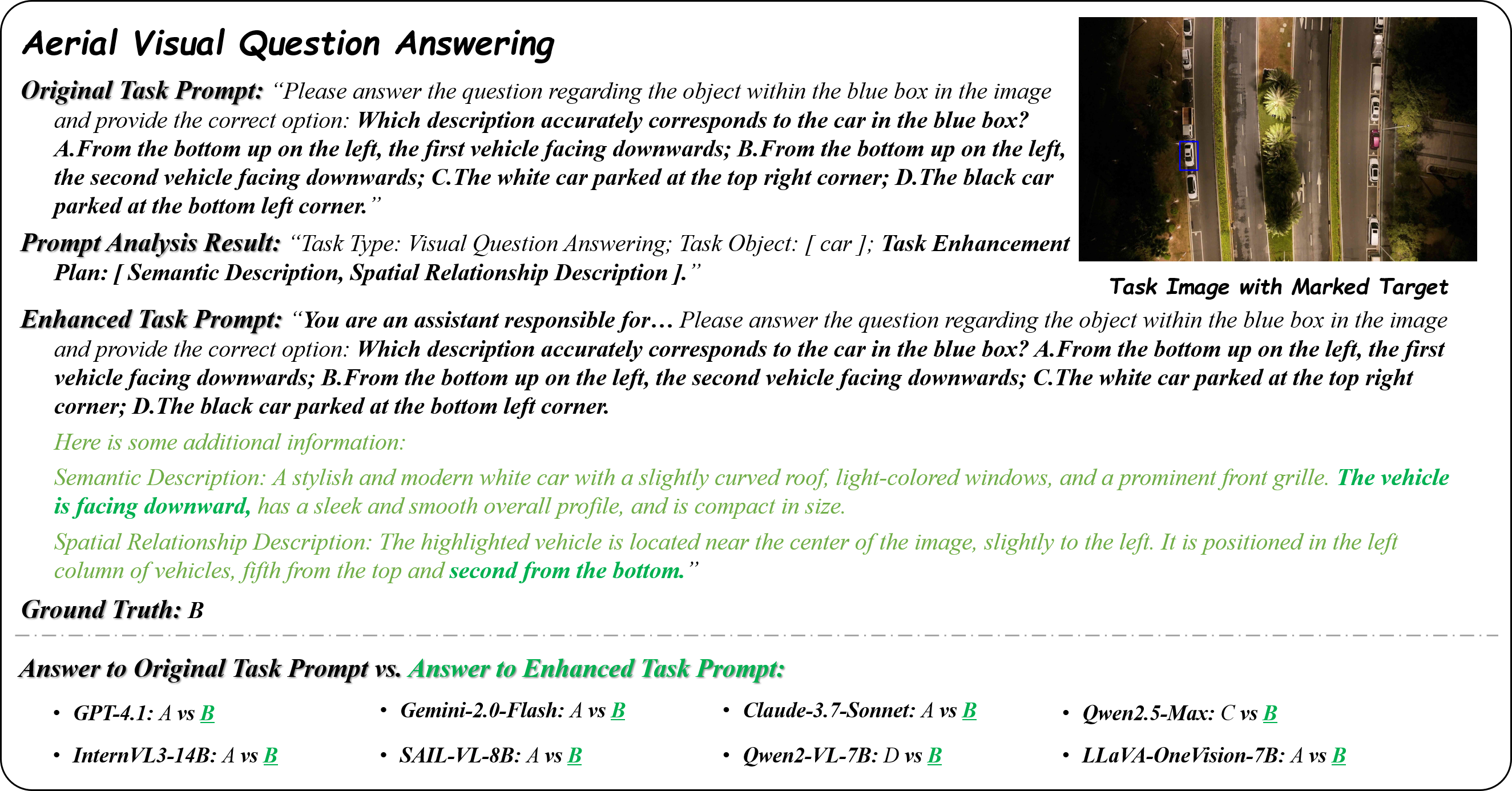}
\vspace{-1mm}
\caption{Illustration of the Aerial Visual Question Answering task guided by prompt enhancement. Similar to the Aerial Visual Reasoning task, the enhanced prompts improve the accuracy of image information perception for most tested models.}
\vspace{-2mm} 
\label{fig:VQA_show}
\end{figure*}

\begin{figure*}[htbp]
\centering
\includegraphics[width=1.0\linewidth]{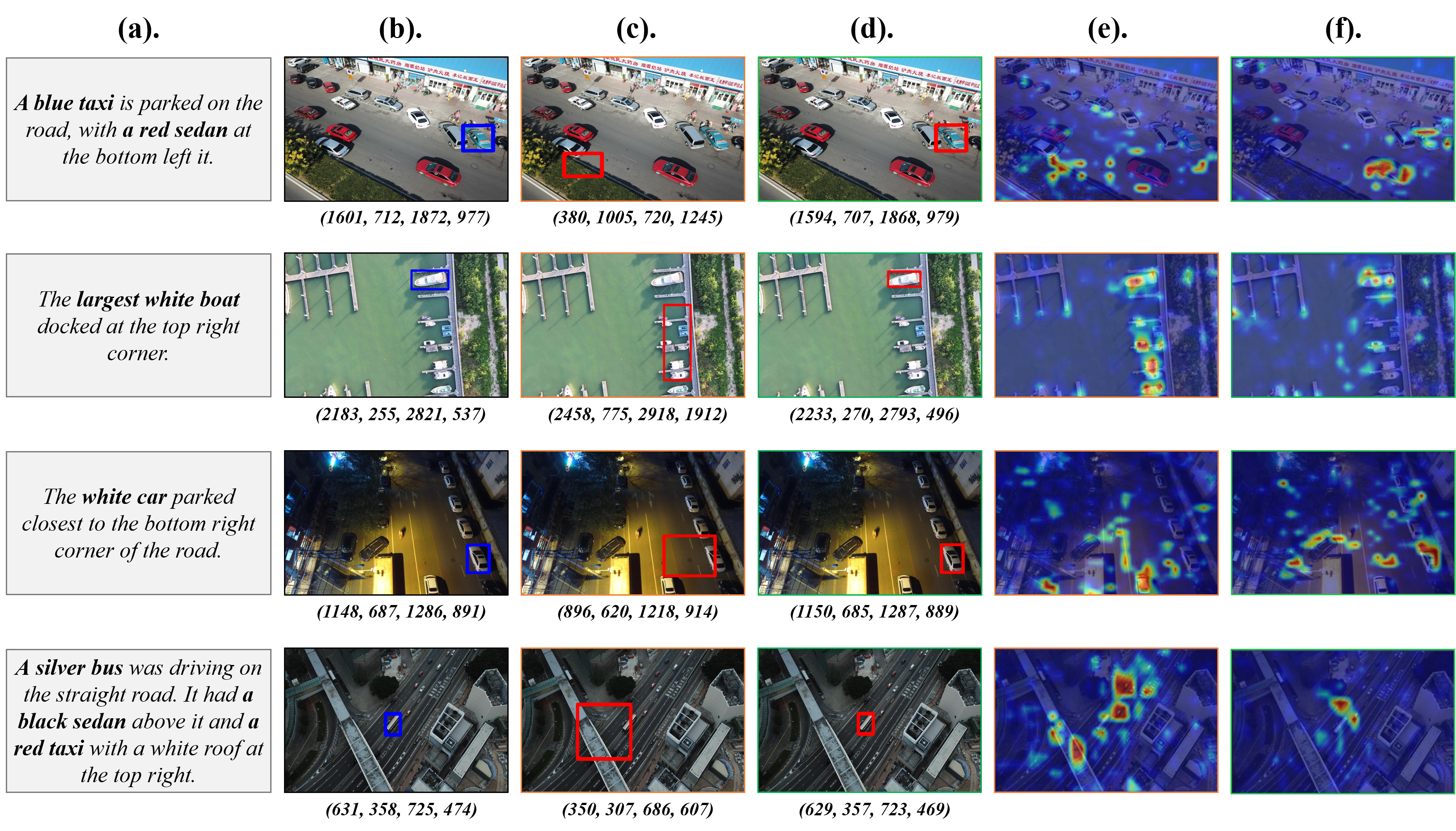}
\caption{Heatmap visualization of the Aerial Visual Grounding task on the LLaVA-1.5-7B model. (a) Task Description; (b) UAV Image with Ground Truth; (c) UAV Image with Answer to Original Task Prompt; (d) UAV Image with Answer to Enhanced Task Prompt; (e) Heatmap to Original Task Prompt; (f) Heatmap to Enhanced Task Prompt. The information below (b) represents the ground-truth answer, while the information below (c) and (d) shows the model’s responses to the corresponding task prompts.}
\label{fig:VG_heatmap}
\end{figure*}

\begin{figure*}[htbp]
\centering
\includegraphics[width=1.0\linewidth]{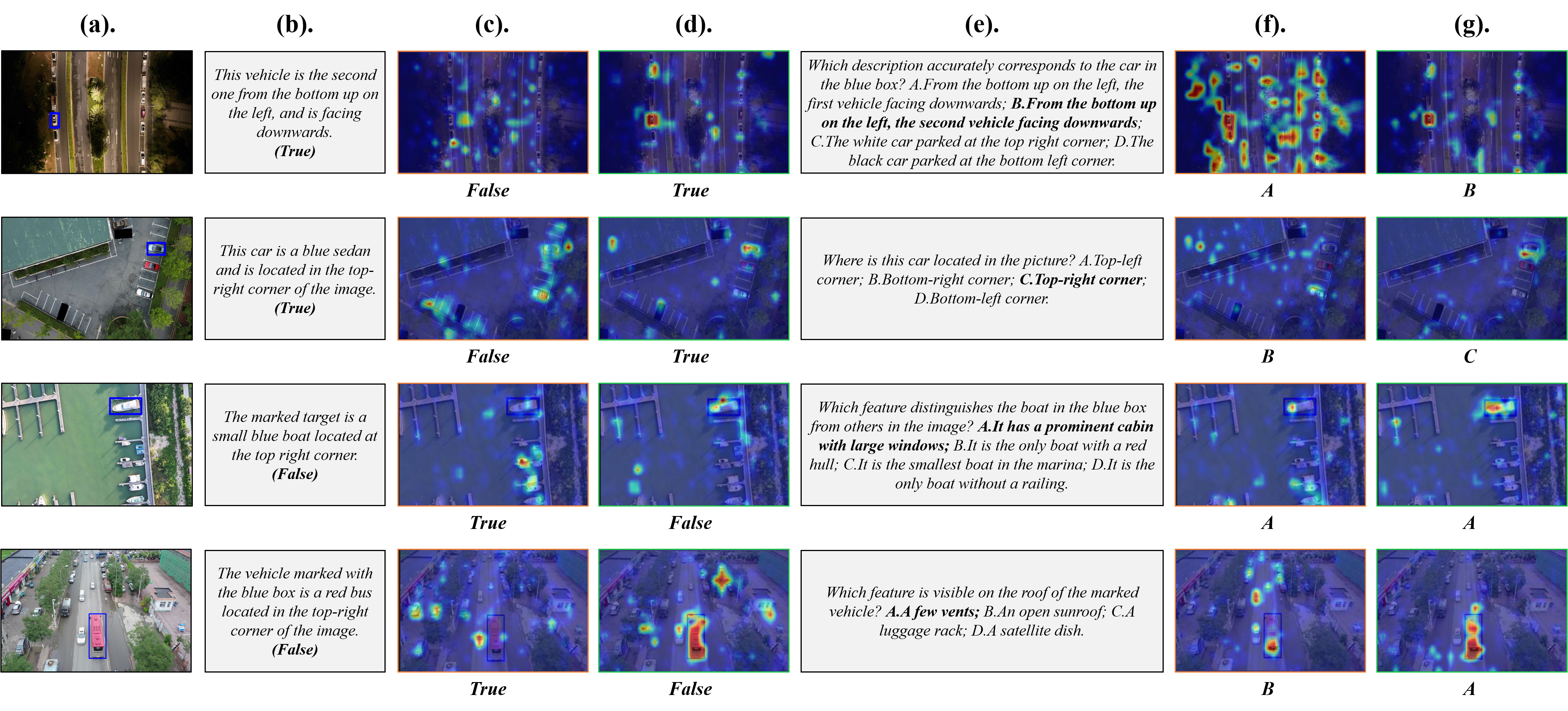}
\caption{Heatmap visualization of the Aerial Visual Reasoning and Aerial Visual Question Answering tasks on the LLaVA-1.5-7B model. (a) Task UAV Image; (b) VR Task Description (the information in brackets indicates the correct answer); (c) Heatmap to Original VR Task Prompt; (d) Heatmap to Enhanced VR Task Prompt; (e) VQA Task Description (the bolded option indicates the correct answer); (f) Heatmap to Original VQA Task Prompt; (g) Heatmap to Enhanced VQA Task Prompt.}
\label{fig:VRVQA_heatmap}
\end{figure*}

\section{Experiments}
\label{sec:experiment}
To systematically evaluate AerialVP, we conduct quantitative and qualitative experiments on the AerialSense benchmark. The experiments are organized around the following research questions:

\begin{itemize}
    \item \textbf{Q1. Overall Effectiveness:} Can AerialVP consistently improve UAV image perception across different tasks and backend VLMs in a training-free manner?
    \item \textbf{Q2. Task-Relevant Visual Guidance:} Do enhanced prompts guide VLMs toward task-relevant visual regions and facilitate task-conditioned image-text alignment?
    \item \textbf{Q3. Comparison with the Model-Centric Framework:} How does the proposed Prompt-Centric Framework compare with LoRA-based Model-Centric fine-tuning under the evaluated settings?
    \item \textbf{Q4. Contribution of Enhancement Dimensions:} What are the individual and combined contributions of semantic, spatial-position, and spatial-relationship information to AerialVP?
\end{itemize}

\subsection{Implementation Details}
\begin{enumerate}
    \item \textbf{Evaluation Models:} To evaluate the general applicability of AerialVP across different VLM architectures, we select representative proprietary and open-source models. The proprietary models include GPT-4o \cite{cite67}, GPT-4.1 \cite{cite68}, Gemini-2.0-Flash \cite{cite69}, Claude-3.7-Sonnet \cite{cite70}, and Qwen2.5-Max \cite{cite71}, while the open-source models comprise InternVL3-8B/14B \cite{cite72}, SAIL-VL-8B \cite{cite73}, Qwen2-VL-7B \cite{cite74}, LLaVA-OneVision-7B \cite{cite55}, and LLaVA-1.5-7B \cite{cite54}. Unless otherwise specified, the main experiments adopt a fixed AerialVP configuration to ensure reproducibility. Specifically, GPT-4.1 serves as the Task Engine, while the Tool Repository employs fixed implementations: GPT-4o for Prompt Analysis Tools, DAM \cite{cite53} for Semantic Description, YOLO-World-X \cite{cite51} for Spatial Position Description, and SpaceOm \cite{cite56} for Spatial Relationship Description.
    
    \item \textbf{Evaluation Metrics:} For the VR and VQA tasks, where each sample has a discrete ground-truth answer, Accuracy is adopted as the primary evaluation metric. For the VG task, we compute the Intersection over Union (IoU) between the predicted and ground-truth bounding boxes. Predictions with an IoU of at least $\beta$ are considered correct, and both Accuracy and Mean IoU are reported. In our experiments, $\beta$ is set to 0.5.
    
    \item \textbf{Experimental Settings:} We consider two evaluation configurations. In the Baseline Configuration, the backend VLM directly performs the perception task using the original task prompt and UAV image. In the Prompt-Enhanced Configuration, AerialVP first enhances the original prompt with task-relevant information, after which the enhanced prompt and the same UAV image are provided to the backend VLM. In addition to quantitative evaluation, attention heatmaps \cite{cite75} are used to qualitatively examine how prompt enhancement affects task-relevant visual attention.
    
    \item \textbf{Comparison with Model-Centric Fine-Tuning Frameworks:} To compare AerialVP with the Model-Centric Framework, we construct an instruction-tuning dataset based on MM-UAVBench \cite{cite76}. The dataset contains 3,125 QA samples covering 10 representative UAV perception tasks, including Class Agnostic Counting, Cross Object Reasoning, Environment State Classification, Ground Target Planning, Orientation Classification, Referring Expression Counting, Scene Classification, Urban OCR, Scene Attribute Understanding, and Scene Damage Assessment. Based on this dataset, we performed LoRA fine-tuning on four open-source VLMs: InternVL3-8B, InternVL3-14B, LLaVA-1.5-7B, and Qwen2-VL-7B. The resulting fine-tuned models are subsequently evaluated on AerialSense, allowing comparison between LoRA-based parameter adaptation and prompt-based adaptation under the evaluated setting. For controlled comparison, we adopt the same LoRA training configuration across the four models. Specifically, LoRA is applied with rank $r=8$ and scaling parameter $\alpha=16$, while the vision encoder remains frozen. Models are trained for 10 epochs using a cosine learning-rate scheduler with an initial learning rate of $5.0 \times 10^{-5}$.
\end{enumerate}

\subsection{Overall Performance on UAV Perception Tasks}
\label{result:sta}
As shown in \textcolor{blue}{Table \ref{tab:VG_acc}}, we present the performance comparison of various VLMs under two configurations for the Aerial VG task. \textcolor{blue}{Table \ref{tab:VRVQA_acc}} further reports the performance comparison on Aerial VR and Aerial VQA. For each model, the two rows correspond to the original prompt and the enhanced prompt configurations, respectively.

Across all three UAV perception tasks, AerialVP consistently improves the performance of both proprietary and open-source VLMs, demonstrating the effectiveness and broad applicability of prompt enhancement. As reported in \textcolor{blue}{Table \ref{tab:VG_acc}}, AerialVP achieves particularly substantial improvements on the VG task, where precise target localization strongly depends on fine-grained image--text alignment. Most proprietary models obtain considerable gains in both Accuracy and Mean IoU. For example, GPT-4o improves from 2.04\% to 45.05\% in Accuracy, while Claude-3.7-Sonnet increases from 8.80\% to 44.93\%. Similar improvements are observed among open-source models, where Qwen2-VL-7B achieves an Accuracy increase from 17.94\% to 37.94\%. These results indicate that enhanced prompts effectively reduce the ambiguity caused by insufficient task guidance in complex UAV scenes. As illustrated in \textcolor{blue}{Fig. \ref{fig:VG_show}}, AerialVP incorporates semantic attributes, spatial positions, and inter-object relationships into the original prompts. Semantic and relational descriptions help models identify task-relevant targets among visually similar objects, while spatial position information provides explicit localization references. Together, these complementary cues provide more explicit guidance for task-conditioned image--text alignment, improving grounding performance in cluttered environments.

\textcolor{blue}{Table \ref{tab:VRVQA_acc}} further demonstrates the effectiveness of AerialVP on VR and VQA tasks. In the VR task, proprietary models achieve accuracy improvements ranging from 13.45 to 29.99 percentage points. For instance, GPT-4.1 improves from 70.40\% to 90.32\%, while open-source models exhibit similar positive trends, with Qwen2-VL-7B increasing from 56.40\% to 85.60\%. Consistent improvements are also observed in VQA. These results demonstrate that prompt enhancement is not limited to localization tasks but also improves performance on higher-level semantic understanding and reasoning tasks. As shown in \textcolor{blue}{Fig. \ref{fig:VR_show}} and \textcolor{blue}{Fig. \ref{fig:VQA_show}}, AerialVP enriches VR and VQA prompts primarily through semantic and spatial relationship information. Semantic information characterizes target appearance and categorical attributes, such as vehicle color or building structure, whereas spatial-relationship information describes relative directions, distances, and neighborhood relations between objects. By integrating these complementary cues, enhanced prompts provide stronger contextual guidance for VLM reasoning, enabling more accurate interpretation of complex UAV scenes.

The experimental results provide three key observations:

\textbf{First, AerialVP consistently improves diverse VLMs across different model architectures.} As shown in \textcolor{blue}{Table \ref{tab:VG_acc}} and \textcolor{blue}{Table \ref{tab:VRVQA_acc}}, both proprietary models (e.g., GPT-4o and Claude-3.7-Sonnet) and open-source models (e.g., InternVL3-8B and Qwen2-VL-7B) benefit from prompt enhancement. This consistency indicates that the effectiveness of AerialVP is not restricted to a specific backend VLM.
    
\textbf{Second, particularly large improvements are observed in Visual Grounding, a spatially demanding task that requires precise target localization.} UAV imagery often contains dense targets, scale variations, and complex spatial layouts. By introducing object-level positions together with semantic and relational information, AerialVP provides more explicit guidance for target localization and discrimination.
    
\textbf{Third, enhanced prompts also improve performance on semantic reasoning tasks by providing richer task-relevant context.} In VR and VQA, semantic and relational descriptions help models form more complete interpretations of target objects and their spatial relationships. As shown in \textcolor{blue}{Fig. \ref{fig:VR_show}} and \textcolor{blue}{Fig. \ref{fig:VQA_show}}, the resulting prompts provide additional information required for reasoning in multi-object UAV scenes.

In summary, these results provide a clear answer to Q1: task-adaptive prompt enhancement consistently improves the perception performance of diverse VLMs across all three AerialSense tasks without task-specific parameter updates to the backend models. By incorporating task-relevant semantic and spatial information, AerialVP provides more informative inference context for localization, visual understanding, and reasoning.

\subsection{Qualitative Analysis Based on Attention Heatmaps}
In addition to quantitative evaluations, we conduct attention heatmap analysis using LLaVA-1.5-7B across Aerial VG, VR, and VQA tasks. This analysis aims to evaluate Q2, specifically whether enhanced prompts can guide the model toward task-relevant visual regions and facilitate task-conditioned image--text alignment.

\textbf{First, the results demonstrate that enhanced task prompts effectively improve the model's localization of task-relevant regions across different UAV perception tasks.} In Aerial VG (\textcolor{blue}{Fig. \ref{fig:VG_heatmap}}), the model exhibits scattered attention over irrelevant regions under the original prompts and fails to consistently focus on the target objects. For instance, when locating the \textit{“blue taxi”} and its reference object \textit{“red sedan”}, the model distributes attention among multiple visually similar vehicles under the original prompt. In contrast, enhanced prompts enable the model to concentrate attention on the correct target and reference objects, producing more precise region correspondence and stronger image--text alignment. Similar improvements are observed in Aerial VR and VQA tasks (\textcolor{blue}{Fig. \ref{fig:VRVQA_heatmap}}), where enhanced prompts guide the model to focus on critical objects, such as the \textit{“red bus”} and \textit{“small blue boat”}, demonstrating more stable task-oriented visual perception.

\textbf{Second, enhanced prompts improve the model's ability to capture contextual objects and spatial cues required for complex reasoning.} In Aerial VG, the model not only attends to the primary target (\textit{“blue taxi”}) but also identifies the auxiliary reference object (\textit{“red sedan”}), which is essential for resolving localization ambiguity in dense aerial scenes. Similar patterns are observed in Aerial VR and VQA tasks, where semantic and spatial-relationship information encourages the model to attend to contextual anchors, including neighboring objects and relative spatial relationships. By incorporating these auxiliary visual cues, enhanced prompts provide richer contextual guidance, enabling more accurate object disambiguation and spatial reasoning.

Overall, the heatmap analysis provides qualitative evidence supporting Q2: enhanced task prompts can redirect VLM attention toward task-relevant regions and contextual references, facilitating more focused image--text alignment during inference. These visualizations complement the quantitative results by providing an interpretable view of how prompt enhancement influences the model's visual attention.

\begin{table*}[ht]
	\centering
	\caption{Comparative Evaluation of Prompt Enhancement and Instruction Fine-Tuning}
	\label{tab:comparative_evaluation}
	\renewcommand{\arraystretch}{1.3}
	\begin{tabular}{M{3.0cm} M{3.0cm} C{2.0cm} C{2.0cm} C{2.0cm} C{2.0cm}}
		\toprule
		\multirow{2}{*}{\textbf{Task Model}} & 
		\multirow{2}{*}{\textbf{Enhancement Strategy}} & 
		\multicolumn{2}{c}{\textbf{VG}} & 
		\multicolumn{1}{c}{\textbf{VQA}} & 
		\multicolumn{1}{c}{\textbf{VR}} \\
		\cmidrule(lr){3-6}
		& & \textbf{Acc} & \textbf{Mean IoU} & \textbf{Acc} & \textbf{Acc} \\
		\midrule
		\multirow{3}{*}{InternVL3-8B} & / & 2.40\% & 7.38\% & 74.91\% & 56.60\% \\
		& Fine-tuning & 6.04\% & 12.52\% & 74.84\% & \textbf{84.12\%} \\
		& Prompt & \textbf{14.21\%} & \textbf{14.50\%} & \textbf{95.40\%} & 78.85\% \\
		\hline
		\multirow{3}{*}{InternVL3-14B} & / & 11.62\% & 18.36\% & 74.93\% & 63.22\% \\
		& Fine-tuning & 20.11\% & \textbf{25.97\%} & 74.78\% & 70.14\% \\
		& Prompt & \textbf{26.59\%} & 24.48\% & \textbf{96.66\%} & \textbf{80.30\%} \\
		\hline
		\multirow{3}{*}{Qwen2-VL-7B} & / & 17.94\% & 19.43\% & 69.76\% & 56.40\% \\
		& Fine-tuning & 17.24\% & 19.09\% & 66.09\% & 77.42\% \\
		& Prompt & \textbf{37.94\%} & \textbf{34.79\%} & \textbf{93.74\%} & \textbf{85.60\%} \\
		\hline
		\multirow{3}{*}{LLaVA-1.5-7B} & / & 7.60\% & 13.57\% & 51.64\% & 53.89\% \\
		& Fine-tuning & 6.19\% & 13.66\% & 50.61\% & 61.15\% \\
		& Prompt & \textbf{22.88\%} & \textbf{21.61\%} & \textbf{90.06\%} & \textbf{90.44\%} \\
		\bottomrule
	\end{tabular}
\end{table*}

\begingroup
\newcolumntype{M}[1]{>{\raggedright\arraybackslash}m{#1}}
\newcolumntype{C}[1]{>{\centering\arraybackslash}m{#1}}
\renewcommand{\arraystretch}{1.4} 

\begin{table*}[ht]
	\centering
	\caption{Ablation Study on Aerial Visual Grounding tasks}
	\label{tab:ablation_VG}
	\begin{tabular}{M{3.0cm} C{2.5cm} C{2.5cm} C{2.5cm} C{2.0cm} C{2.0cm}}
		\toprule
		Task Model & Semantic Enhancement & Spatial Position Enhancement & Spatial Relationship Enhancement & Acc & Mean IoU \\
		\midrule
		\cellcolor{lightgray}\textit{Proprietary Models} & \cellcolor{lightgray} & \cellcolor{lightgray} & \cellcolor{lightgray} & \cellcolor{lightgray} & \cellcolor{lightgray} \\
		\hline
		\multirow{7}{*}{Qwen2.5-Max} & $\times$ & $\times$ & $\times$ & 1.22\% & 7.17\% \\
		& $\checkmark$ & $\times$ & $\times$ & 7.18\% & 6.34\% \\
		& $\times$ & $\checkmark$ & $\times$ & 22.40\% & 19.85\% \\
		& $\times$ & $\times$ & $\checkmark$ & 0.07\% & 0.14\% \\
		& $\checkmark$ & $\checkmark$ & $\times$ & 38.33\% & 33.88\% \\
		& $\times$ & $\checkmark$ & $\checkmark$ & 24.93\% & 22.07\% \\
		& $\checkmark$ & $\checkmark$ & $\checkmark$ & 43.87\% & 39.87\% \\
		\midrule
		\cellcolor{lightgray}\textit{Open-Source Models} & \cellcolor{lightgray} & \cellcolor{lightgray} & \cellcolor{lightgray} & \cellcolor{lightgray} & \cellcolor{lightgray} \\
		\hline
		\multirow{7}{*}{Qwen2-VL-7B} & $\times$ & $\times$ & $\times$ & 17.94\% & 19.43\% \\
		& $\checkmark$ & $\times$ & $\times$ & 23.73\% & 21.04\% \\
		& $\times$ & $\checkmark$ & $\times$ & 27.70\% & 24.64\% \\
		& $\times$ & $\times$ & $\checkmark$ & 4.92\% & 6.80\% \\
		& $\checkmark$ & $\checkmark$ & $\times$ & 32.17\% & 28.57\% \\
		& $\times$ & $\checkmark$ & $\checkmark$ & 21.57\% & 19.37\% \\
		& $\checkmark$ & $\checkmark$ & $\checkmark$ & 37.94\% & 34.79\% \\
		\bottomrule
	\end{tabular}
\end{table*}

\begin{table*}[ht]
	\centering
	\caption{Ablation Study on Aerial Visual Reasoning (VR) and Aerial Visual Question Answering (VQA) tasks}
	\label{tab:ablation_VRVQA}
	\begin{tabular}{M{3.0cm} C{3.0cm} C{4.4cm} C{2.0cm} C{2.0cm}}
		\toprule
		Task Model & Semantic Enhancement & Spatial Relationship Enhancement & Acc(VR) & Acc(VQA) \\
		\midrule
		\cellcolor{lightgray}\textit{Proprietary Models} & \cellcolor{lightgray} & \cellcolor{lightgray} & \cellcolor{lightgray} & \cellcolor{lightgray} \\
		\hline
		\multirow{4}{*}{Qwen2.5-Max} & $\times$ & $\times$ & 77.03\% & 65.95\% \\
		& $\checkmark$ & $\times$ & 87.29\% & 73.66\% \\
		& $\times$ & $\checkmark$ & 82.71\% & 70.66\% \\
		& $\checkmark$ & $\checkmark$ & 90.48\% & 77.54\% \\
		\midrule
		\cellcolor{lightgray}\textit{Open-Source Models} & \cellcolor{lightgray} & \cellcolor{lightgray} & \cellcolor{lightgray} & \cellcolor{lightgray} \\
		\hline
		\multirow{4}{*}{Qwen2-VL-7B} & $\times$ & $\times$ & 56.40\% & 69.76\% \\
		& $\checkmark$ & $\times$ & 76.17\% & 88.06\% \\
		& $\times$ & $\checkmark$ & 65.69\% & 76.57\% \\
		& $\checkmark$ & $\checkmark$ & 85.60\% & 93.74\% \\
		\bottomrule
	\end{tabular}
\end{table*}
\endgroup

\subsection{Comparative Analysis with Model-Centric Frameworks}
As shown in \textcolor{blue}{Table \ref{tab:comparative_evaluation}}, we compare the two adaptation paradigms using LoRA-based instruction fine-tuning as a representative Model-Centric Framework and AerialVP as the implementation of the Prompt-Centric Framework. The comparison is conducted on four open-source VLMs across VG, VQA, and VR tasks. For each model, the three rows report the original baseline (denoted by \textit{“/”}), LoRA-based fine-tuning, and AerialVP-based prompt enhancement, respectively.

Overall, AerialVP achieves competitive and often higher performance than LoRA-based fine-tuning, although their relative advantages vary across tasks and models. On VG, AerialVP achieves higher Accuracy than fine-tuning for all four models and higher Mean IoU in three of the four cases; only InternVL3-14B obtains a slightly higher Mean IoU after fine-tuning. On VQA, fine-tuning provides little benefit and even slightly degrades the baseline performance for all four evaluated models, whereas AerialVP produces substantial improvements consistently. On VR, both strategies are effective, but their relative performance is model-dependent: fine-tuning performs better for InternVL3-8B, while AerialVP achieves higher Accuracy for InternVL3-14B, Qwen2-VL-7B, and LLaVA-1.5-7B. These results show that prompt enhancement is particularly effective in VG and VQA, while the relative advantage over fine-tuning varies more noticeably in VR.

These results reveal two main observations regarding the two adaptation strategies:

\textbf{First, the Prompt-Centric Framework exhibits more consistent performance gains across the evaluated models and tasks.} LoRA-based fine-tuning improves several model-task combinations, but its gains vary substantially and are negligible or negative in some settings, particularly on VQA and several VG configurations. In contrast, AerialVP yields positive improvements over the original baseline in every reported model-task setting and evaluation metric, although the magnitude of the gains varies across models and tasks. This consistency indicates that prompt-side adaptation provides a stable means of improving UAV perception across heterogeneous VLM architectures.

\textbf{Second, the two paradigms exhibit complementary performance profiles rather than universal superiority of one over the other.} AerialVP achieves considerably larger gains in many VG and VQA settings and consistently improves the original models across all evaluated tasks. However, its gains are lower than those of fine-tuning in several specific settings, such as VR for InternVL3-8B and Mean IoU on VG for InternVL3-14B. These observations suggest that parameter adaptation and prompt enhancement offer different advantages, with their relative effectiveness depending on the backend model and task characteristics.

In summary, these results answer Q3: the Prompt-Centric framework achieves competitive performance with LoRA-based Model-Centric fine-tuning and outperforms it in many evaluated settings, while delivering more consistent improvements across models and tasks. Meanwhile, fine-tuning remains advantageous in several specific settings. AerialVP therefore provides a competitive training-free alternative for UAV perception tasks.

\subsection{Ablation Study}
To investigate the contribution of different enhancement components in AerialVP, we conduct an ablation study by controlling the activation of enhancement tools through the Task Enhancement Plan. Using proprietary Qwen2.5-Max and open-source Qwen2-VL-7B as test models, we evaluate the individual and combined effects of spatial-position, semantic, and spatial-relationship enhancement. The detailed results are presented in \textcolor{blue}{Table \ref{tab:ablation_VG}} and \textcolor{blue}{Table \ref{tab:ablation_VRVQA}}.

For the Aerial VG task (\textcolor{blue}{Table \ref{tab:ablation_VG}}), different enhancement types exhibit distinct contributions to localization performance. Spatial-position enhancement provides the strongest individual improvement, demonstrating the importance of explicit positional information for resolving target-localization ambiguity in complex UAV scenes. Semantic enhancement alone also generally improves performance by providing additional target attributes. In contrast, spatial-relationship enhancement alone provides limited benefits and may even degrade performance. For example, Qwen2.5-Max decreases from 1.22\% to 0.07\% in Accuracy and from 7.17\% to 0.14\% in Mean IoU when only spatial-relationship information is introduced. A similar trend is observed for Qwen2-VL-7B. This suggests that relational information is difficult to utilize effectively in isolation and becomes more beneficial when combined with complementary positional and semantic cues.

When spatial-position and semantic enhancement are jointly applied, performance improves substantially for both models, indicating that semantic attributes complement positional constraints in target discrimination. Combining all three enhancement types achieves the best overall performance, further demonstrating that spatial-relationship information becomes beneficial when supported by positional and semantic context.

For Aerial VR and VQA tasks (\textcolor{blue}{Table \ref{tab:ablation_VRVQA}}), semantic and spatial-relationship enhancements both provide positive contributions because these tasks rely more strongly on object understanding and relational reasoning. For Qwen2.5-Max, semantic enhancement improves VR Accuracy from 77.03\% to 87.29\% and VQA Accuracy from 65.95\% to 73.66\%, while spatial-relationship enhancement increases them to 82.71\% and 70.66\%, respectively. Similar trends are observed with Qwen2-VL-7B. Combining both components consistently achieves the best performance, demonstrating their complementary roles in semantic-spatial reasoning.

In summary, these ablation results answer Q4 by revealing the distinct and complementary roles of different enhancement dimensions. Spatial-position information is particularly important for grounding, semantic information improves target discrimination and understanding, and spatial-relationship information supports relational reasoning. Importantly, the results demonstrate that prompt enhancement should be selectively performed according to task requirements rather than simply incorporating more information. In particular, isolated relational cues may introduce additional ambiguity when lacking sufficient semantic or positional context, whereas appropriately combined enhancement dimensions provide more effective task guidance.


\begingroup
\newcolumntype{M}[1]{>{\raggedright\arraybackslash}m{#1}}
\newcolumntype{C}[1]{>{\centering\arraybackslash}m{#1}}
\renewcommand{\arraystretch}{1.4}

\begin{table*}[ht]
	\centering
	\caption{Performance Improvements of Different VLMs on UAV Perception Tasks under Enhanced Task Prompts}
	\label{tab:prompt_improvement}
	\begin{tabular}{M{3.0cm} C{2.8cm} C{2.8cm} C{2.8cm} C{2.8cm}}
		\toprule
		\multicolumn{1}{l}{Model} & \multicolumn{1}{c}{Acc (VG)} & \multicolumn{1}{c}{Mean IoU (VG)} & \multicolumn{1}{c}{Acc (VR)} & \multicolumn{1}{c}{Acc (VQA)} \\
		\midrule
		\cellcolor{lightgray}\textit{Proprietary Models} 
		& \cellcolor{lightgray} & \cellcolor{lightgray} 
		& \cellcolor{lightgray} & \cellcolor{lightgray} \\
		\hline
		GPT-4o      & +43.01\% & +37.57\% & +29.99\% & +20.50\% \\
		GPT-4.1     & +45.68\% & +32.96\% & +19.92\% & +13.79\% \\
		Gemini-2.0-Flash  & +40.92\% & +31.23\% & +14.75\% & +21.53\% \\
		Claude-3.7-Sonnet  & +36.13\% & +28.14\% & +22.70\% & +18.68\% \\
		Qwen2.5-Max & +42.65\% & +32.70\% & +13.45\% & +11.59\% \\
		\midrule
		\cellcolor{lightgray}\textit{Open-Source Models} 
		& \cellcolor{lightgray} & \cellcolor{lightgray} 
		& \cellcolor{lightgray} & \cellcolor{lightgray} \\
		\hline
		InternVL3-8B        & +11.81\% & +7.12\%  & +22.25\% & +20.49\% \\
		InternVL3-14B       & +14.97\% & +6.12\%  & +17.08\% & +21.73\% \\
		SAIL-VL-8B            & +28.08\% & +20.89\% & +27.20\% & +56.92\% \\
		Qwen2-VL-7B          & +20.00\% & +15.36\% & +29.20\% & +23.98\% \\
		LLaVA-OneVision-7B   & +30.02\% & +21.43\% & +14.43\% & +21.11\% \\
		LLaVA-1.5-7B         & +15.28\% & +8.04\%  & +36.55\% & +38.42\% \\
		\bottomrule
	\end{tabular}
\end{table*}
\endgroup

\begingroup
\newcolumntype{M}[1]{>{\raggedright\arraybackslash}m{#1}}
\newcolumntype{C}[1]{>{\centering\arraybackslash}m{#1}}
\renewcommand{\arraystretch}{1.4} 

\begin{table*}[ht]
\centering
\caption{Comparison between Different \textbf{Semantic Tools} on Visual Grounding tasks}
\label{tab:semantic_comparison}
\begin{tabular}{M{2.5cm} C{3.0cm} C{2.0cm} C{2.8cm} C{1.8cm} C{1.8cm}}
\toprule
\multicolumn{1}{l}{Task Model} & \multicolumn{3}{c}{Prompt Enhancement Tool} & \multicolumn{1}{c}{Acc} & \multicolumn{1}{c}{Mean IoU} \\
\cmidrule(lr){2-4}
& \textbf{Semantic} & Spatial Position & Spatial Relationship & & \\
\midrule
\cellcolor{lightgray}\textit{Proprietary Models} & \cellcolor{lightgray} & \cellcolor{lightgray} & \cellcolor{lightgray} & \cellcolor{lightgray} & \cellcolor{lightgray} \\
\hline
\multirow{3}{*}{GPT-4.1} & \textbf{DAM} & YoloWorld-X & SpaceOm & 49.28\% & 44.87\% \\
& \textbf{LLaVA-1.5-7B} & YoloWorld-X & SpaceOm & 44.00\% & 40.56\% \\
& \textbf{LLaVA-OneVision-7B} & YoloWorld-X & SpaceOm & 47.58\% & 43.39\% \\
\hline
\multirow{3}{*}{Gemini-2.0-Flash} & \textbf{DAM} & YoloWorld-X & SpaceOm & 51.09\% & 46.22\% \\
& \textbf{LLaVA-1.5-7B} & YoloWorld-X & SpaceOm & 46.55\% & 42.27\% \\
& \textbf{LLaVA-OneVision-7B} & YoloWorld-X & SpaceOm & 48.74\% & 44.12\% \\
\hline
\multirow{3}{*}{Qwen2.5-Max} & \textbf{DAM} & YoloWorld-X & SpaceOm & 43.87\% & 39.87\% \\
& \textbf{LLaVA-1.5-7B} & YoloWorld-X & SpaceOm & 37.33\% & 34.39\% \\
& \textbf{LLaVA-OneVision-7B} & YoloWorld-X & SpaceOm & 41.45\% & 37.83\% \\
\midrule
\cellcolor{lightgray}\textit{Open-Source Models} & \cellcolor{lightgray} & \cellcolor{lightgray} & \cellcolor{lightgray} & \cellcolor{lightgray} & \cellcolor{lightgray} \\
\hline
\multirow{3}{*}{InternVL3-14B} & \textbf{DAM} & YoloWorld-X & SpaceOm & 26.59\% & 24.48\% \\
& \textbf{LLaVA-1.5-7B} & YoloWorld-X & SpaceOm & 23.43\% & 21.84\% \\
& \textbf{LLaVA-OneVision-7B} & YoloWorld-X & SpaceOm & 24.95\% & 23.16\% \\
\hline
\multirow{3}{*}{SAIL-VL-8B} & \textbf{DAM} & YoloWorld-X & SpaceOm & 34.23\% & 31.82\% \\
& \textbf{LLaVA-1.5-7B} & YoloWorld-X & SpaceOm & 32.72\% & 30.65\% \\
& \textbf{LLaVA-OneVision-7B} & YoloWorld-X & SpaceOm & 33.37\% & 31.20\% \\
\bottomrule
\end{tabular}
\end{table*}

\begin{table*}[ht]
\centering
\caption{Comparison between Different \textbf{Spatial Position Tools} on Visual Grounding tasks}
\label{tab:coordinate_comparison}
\begin{tabular}{M{2.5cm} C{2.0cm} C{3.0cm} C{2.8cm} C{1.8cm} C{1.8cm}}
\toprule
\multicolumn{1}{l}{Task Model} & \multicolumn{3}{c}{Prompt Enhancement Tool} & \multicolumn{1}{c}{Acc} & \multicolumn{1}{c}{Mean IoU} \\
\cmidrule(lr){2-4}
& Semantic & \textbf{Spatial Position} & Spatial Relationship & & \\
\midrule
\cellcolor{lightgray}\textit{Proprietary Models} & \cellcolor{lightgray} & \cellcolor{lightgray} & \cellcolor{lightgray} & \cellcolor{lightgray} & \cellcolor{lightgray} \\
\hline
\multirow{2}{*}{GPT-4.1} & DAM & \textbf{YoloWorld-X} & SpaceOm & 49.28\% & 44.87\% \\
& DAM & \textbf{Grounding-DINO} & SpaceOm & 40.21\% & 37.21\% \\
\hline
\multirow{2}{*}{Gemini-2.0-Flash} & DAM & \textbf{YoloWorld-X} & SpaceOm & 51.09\% & 46.22\% \\
& DAM & \textbf{Grounding-DINO} & SpaceOm & 42.65\% & 38.90\% \\
\hline
\multirow{2}{*}{Qwen2.5-Max} & DAM & \textbf{YoloWorld-X} & SpaceOm & 43.87\% & 39.87\% \\
& DAM & \textbf{Grounding-DINO} & SpaceOm & 35.62\% & 32.89\% \\
\midrule
\cellcolor{lightgray}\textit{Open-Source Models} & \cellcolor{lightgray} & \cellcolor{lightgray} & \cellcolor{lightgray} & \cellcolor{lightgray} & \cellcolor{lightgray} \\
\hline
\multirow{2}{*}{InternVL3-14B} & DAM & \textbf{YoloWorld-X} & SpaceOm & 26.59\% & 24.48\% \\
& DAM & \textbf{Grounding-DINO} & SpaceOm & 21.64\% & 20.18\% \\
\hline
\multirow{2}{*}{SAIL-VL-8B} & DAM & \textbf{YoloWorld-X} & SpaceOm & 34.23\% & 31.82\% \\
& DAM & \textbf{Grounding-DINO} & SpaceOm & 30.78\% & 28.81\% \\
\bottomrule
\end{tabular}
\end{table*}

\begin{table*}[ht]
\centering
\caption{Comparison between Different \textbf{Spatial Relationship Tools} on Visual Grounding tasks}
\label{tab:spatial_comparison}
\begin{tabular}{M{2.5cm} C{2.0cm} C{3.0cm} C{2.8cm} C{1.8cm} C{1.8cm}}
\toprule
\multicolumn{1}{l}{Task Model} & \multicolumn{3}{c}{Prompt Enhancement Tool} & \multicolumn{1}{c}{Acc} & \multicolumn{1}{c}{Mean IoU} \\
\cmidrule(lr){2-4}
& Semantic & Spatial Position & \textbf{Spatial Relationship} & & \\
\midrule
\cellcolor{lightgray}\textit{Proprietary Models} & \cellcolor{lightgray} & \cellcolor{lightgray} & \cellcolor{lightgray} & \cellcolor{lightgray} & \cellcolor{lightgray} \\
\hline
\multirow{3}{*}{GPT-4.1} & DAM & YoloWorld-X & \textbf{SpaceOm} & 49.28\% & 44.87\% \\
& DAM & YoloWorld-X & \textbf{LLaVA-1.5-7B} & 62.00\% & 55.53\% \\
& DAM & YoloWorld-X & \textbf{LLaVA-OneVision-7B} & 58.32\% & 52.64\% \\
\hline
\multirow{3}{*}{Gemini-2.0-Flash} & DAM & YoloWorld-X & \textbf{SpaceOm} & 51.09\% & 46.22\% \\
& DAM & YoloWorld-X & \textbf{LLaVA-1.5-7B} & 59.58\% & 53.48\% \\
& DAM & YoloWorld-X & \textbf{LLaVA-OneVision-7B} & 58.36\% & 52.50\% \\
\hline
\multirow{3}{*}{Qwen2.5-Max} & DAM & YoloWorld-X & \textbf{SpaceOm} & 43.87\% & 39.87\% \\
& DAM & YoloWorld-X & \textbf{LLaVA-1.5-7B} & 49.52\% & 44.75\% \\
& DAM & YoloWorld-X & \textbf{LLaVA-OneVision-7B} & 49.81\% & 44.85\% \\
\midrule
\cellcolor{lightgray}\textit{Open-Source Models} & \cellcolor{lightgray} & \cellcolor{lightgray} & \cellcolor{lightgray} & \cellcolor{lightgray} & \cellcolor{lightgray} \\
\hline
\multirow{3}{*}{InternVL3-14B} & DAM & YoloWorld-X & \textbf{SpaceOm} & 26.59\% & 24.48\% \\
& DAM & YoloWorld-X & \textbf{LLaVA-1.5-7B} & 30.32\% & 27.73\% \\
& DAM & YoloWorld-X & \textbf{LLaVA-OneVision-7B} & 33.37\% & 30.23\% \\
\hline
\multirow{3}{*}{SAIL-VL-8B} & DAM & YoloWorld-X & \textbf{SpaceOm} & 34.23\% & 31.82\% \\
& DAM & YoloWorld-X & \textbf{LLaVA-1.5-7B} & 38.04\% & 35.20\% \\
& DAM & YoloWorld-X & \textbf{LLaVA-OneVision-7B} & 38.15\% & 35.41\% \\
\bottomrule
\end{tabular}
\end{table*}
\endgroup

\begin{figure*}[h]
\centering
\includegraphics[width=0.9\linewidth]{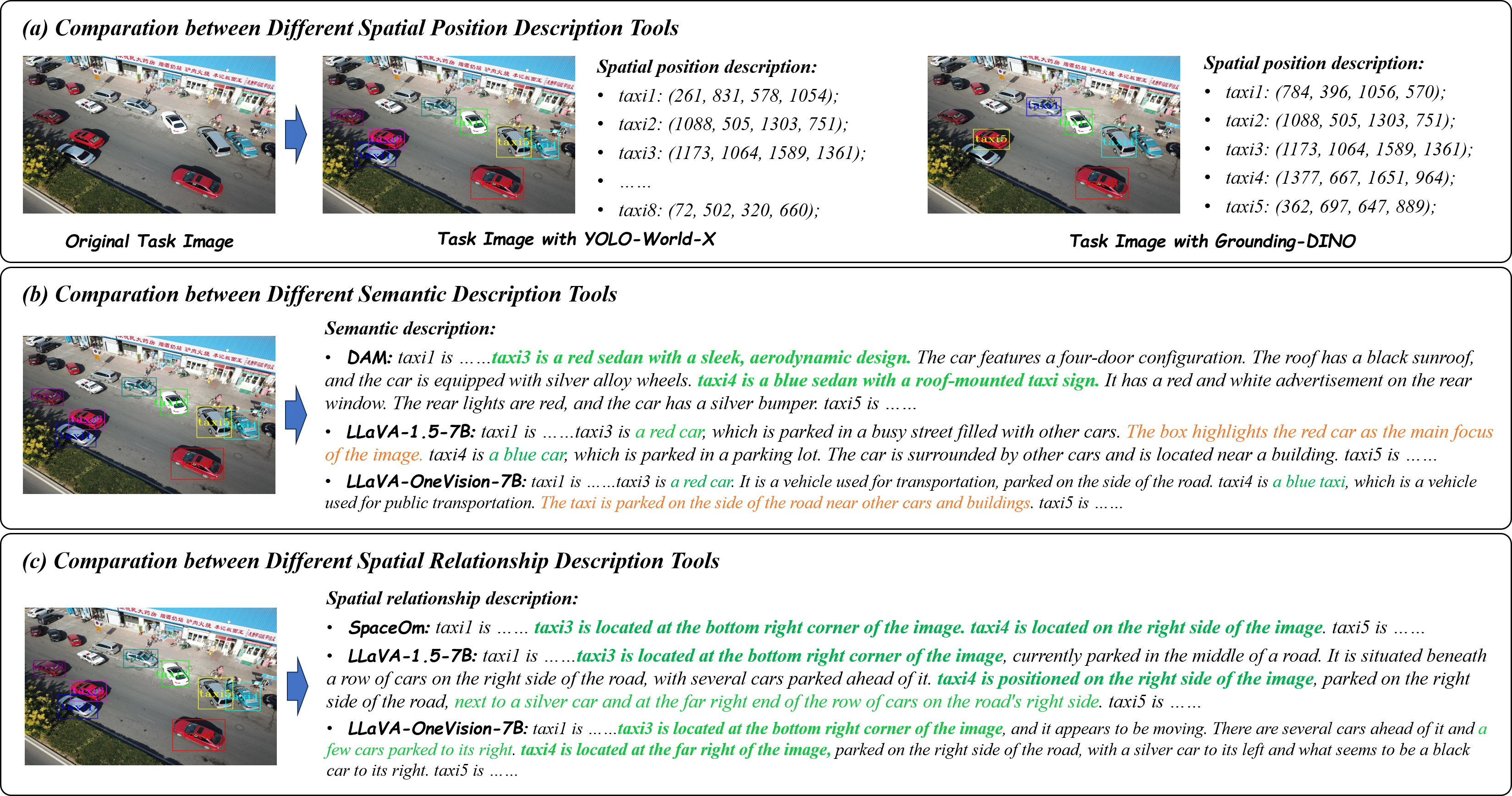}
\vspace{-1mm}
\caption{Description comparison of different enhancement tools in the Aerial Visual Grounding task. For clear visualization of enhancement effects, each subfigure highlights representative targets (e.g., “taxi3” and “taxi4” in (a)) that critically influence task performance.}
\vspace{-2mm} 
\label{fig:compare_VGVRVQA}
\end{figure*}

\section{Discussion}
The experiments in \textcolor{blue}{Section \ref{sec:experiment}} demonstrate the effectiveness of AerialVP across different UAV perception tasks and analyze the contributions of its major enhancement dimensions. In this section, we further investigate two factors that influence prompt-enhancement performance: the characteristics of the backend VLM and the implementations of the Prompt Enhancement Tools. These analyses provide further insights into the generality and extensibility of the proposed Prompt-Centric Framework.
\subsection{Relationship Between Prompt Enhancement and the Reasoning Ability of Different VLMs}
AerialVP enhances the task context provided to the backend VLM, while the final perception and reasoning processes are still performed by the backend model itself. Therefore, different VLMs may exhibit different abilities to interpret and utilize the semantic and spatial information introduced by enhanced prompts. To examine this effect, \textcolor{blue}{Table \ref{tab:prompt_improvement}} summarizes the performance gains achieved by all evaluated VLMs across the three AerialSense tasks based on the results in \textcolor{blue}{Section \ref{result:sta}}.

Overall, all evaluated VLMs benefit from prompt enhancement, but the magnitude of improvement varies considerably across models and tasks. In Aerial VG, proprietary models generally obtain larger gains. For example, GPT-4o, GPT-4.1, Gemini-2.0-Flash, Claude-3.7-Sonnet, and Qwen2.5-Max achieve Accuracy gains of 43.01, 45.68, 40.92, 36.13, and 42.65 percentage points, respectively. In comparison, the corresponding gains among open-source models range from 11.81 to 30.02 percentage points. However, this pattern does not hold uniformly for VR and VQA. Several open-source models achieve gains comparable to or larger than those of proprietary models. For instance, LLaVA-1.5-7B gains 36.55 percentage points on VR, while SAIL-VL-8B gains 56.92 percentage points on VQA. These results indicate that the effectiveness of prompt enhancement varies across model architectures and task types, rather than following a uniform pattern across proprietary and open-source VLMs.

Therefore, these results highlight the interaction between enhanced task guidance and backend VLM reasoning capability: AerialVP provides richer task contexts, while the backend VLM determines how effectively these additional cues are interpreted and utilized for final prediction. Different models may exhibit different sensitivities to semantic descriptions, explicit spatial cues, and relational information, leading to varying improvements across task types.

These observations suggest that future VLM-based perception research should not consider model capability improvement and prompt information enhancement as isolated directions. Instead, jointly optimizing the intrinsic reasoning capability of the model and the quality of task-specific guidance may provide a more effective pathway toward achieving accurate and robust multimodal perception.

\subsection{Impact of Tool Selection on Model Performance Improvement}
The quality of an enhanced prompt is also influenced by the Prompt Enhancement Tools used to extract auxiliary information. Benefiting from the modular design of the Tool Repository, individual tools can be replaced by functionally equivalent implementations without changing the overall AerialVP workflow. To investigate this property, we compare alternative implementations of the spatial-position, semantic-description, and spatial-relationship tools on the Aerial VG task. The results are reported in \textcolor{blue}{Tables \ref{tab:semantic_comparison} - \ref{tab:spatial_comparison}}, with representative examples shown in \textcolor{blue}{Fig. \ref{fig:compare_VGVRVQA}}.

First, different tool implementations have a substantial impact on the effectiveness of prompt enhancement. These performance differences indicate that the quality of auxiliary information generated by the enhancement tools can substantially influence the final perception results. Specifically, as shown in \textcolor{blue}{Table \ref{tab:coordinate_comparison}}, YOLO-World-X consistently achieves higher Accuracy and Mean IoU than Grounding-DINO across the evaluated VLMs, indicating its stronger capability to provide reliable spatial-position information. Similarly, \textcolor{blue}{Table \ref{tab:semantic_comparison}} shows that DAM provides more effective semantic enhancement than the LLaVA-based alternatives, while \textcolor{blue}{Table \ref{tab:spatial_comparison}} demonstrates that the LLaVA-based implementations generally achieve better spatial-relationship enhancement than SpaceOm. These results show that different implementations of the same functional tool can vary substantially in the quality of the auxiliary information they provide, leading to different prompt-enhancement performance.

Second, the performance differences among tool implementations are closely related to the granularity, completeness, and contextual informativeness of the generated auxiliary information. As illustrated in \textcolor{blue}{Fig. \ref{fig:compare_VGVRVQA} (a)}, YOLO-World-X provides more comprehensive target localization results, whereas Grounding-DINO may miss small or partially occluded targets, leading to incomplete spatial cues. For semantic enhancement (\textcolor{blue}{Fig. \ref{fig:compare_VGVRVQA} (b)}), DAM generates fine-grained descriptions involving object categories, visual attributes, and contextual information, while the LLaVA-based tools tend to produce more general descriptions with fewer discriminative details. For spatial-relationship enhancement (\textcolor{blue}{Fig. \ref{fig:compare_VGVRVQA} (c)}), the LLaVA-based tools generate more context-aware relational descriptions, such as relative directions and neighborhood relations, whereas SpaceOm mainly provides geometric relationship information with weaker semantic context. These observations suggest that fine-grained, complete, and context-aware auxiliary information can provide more effective task guidance for image--text alignment and reasoning.

Overall, these results demonstrate that AerialVP is sensitive to the quality of its underlying tool implementations. More accurate localization, more discriminative semantic descriptions, and more informative relational cues generally lead to better prompt-enhancement performance. But the modular design of the Tool Repository allows individual tools to be flexibly replaced or extended without modifying the overall workflow, enabling the integration of more capable implementations for different information dimensions.

Looking ahead, we envision that VLM-based perception systems may evolve beyond fixed-capability architectures toward more flexible and extensible frameworks, where specialized tools can be dynamically incorporated to accommodate increasingly diverse and complex perception requirements.


\section{Conclusion}
This paper provides a new perspective on VLM-based UAV image perception by analyzing its challenges from the perspective of task-conditioned image–text alignment. We reveal that the perception difficulty in complex UAV scenarios is closely related to the mismatch between fine-grained visual complexity and limited textual guidance, which leads to ambiguous associations between textual descriptions and visual targets. Inspired by the role of discriminative guidance in human visual perception, we propose a training-free Prompt-Centric paradigm that improves alignment by enriching the textual references available during inference rather than optimizing model parameters.

Based on this paradigm, we develop AerialVP, a task-adaptive prompt-enhancement agent that dynamically determines task-specific information requirements and incorporates image-derived semantic and spatial cues into task prompts. Compared with conventional fine-tuning-based adaptation, AerialVP improves UAV perception without modifying backend VLM parameters and enables flexible application across both open-source and proprietary VLMs. We further construct AerialSense, a unified multi-task benchmark with diverse targets, scenes, resolutions, and imaging conditions to support systematic evaluation of VLM-based UAV perception. Extensive experiments demonstrate that AerialVP consistently improves diverse VLMs across multiple perception tasks and achieves competitive performance against LoRA-based instruction fine-tuning, validating the effectiveness of prompt-centric adaptation.

Nevertheless, this work still has several limitations. First, the performance of AerialVP depends on the quality of the underlying Prompt Enhancement Tools, and more powerful or specialized tools may further improve the generated task guidance. Second, the current benchmark and evaluation focus on static UAV images, while real-world UAV applications often involve dynamic environments and continuous perception. Future work will explore more adaptive tool-selection strategies and extend prompt-centric adaptation toward broader UAV perception scenarios, including dynamic and embodied perception tasks.


\bibliographystyle{IEEEtran}
\bibliography{IEEEabrv,main}

\end{document}